\pdfoutput=1
\documentclass{article}

\usepackage{microtype}
\usepackage{graphicx}
\usepackage{subcaption}
\usepackage{booktabs} 

\usepackage{hyperref}

\usepackage[accepted]{icml2023}

\usepackage{amsmath}
\usepackage{amssymb}
\usepackage{mathtools}
\usepackage{amsthm}
\usepackage{multirow}
\usepackage{makecell}
\usepackage{nicefrac}

\usepackage{color}
\usepackage{xcolor}
\usepackage{xspace}

\newcommand{\update}[1]{#1}

\newcommand{\nameshort}{OMS-DPM}

\usepackage[capitalize,noabbrev,nameinlink]{cleveref}

\crefname{section}{Sec.}{Secs.}
\Crefname{section}{Sec.}{Secs.}
\Crefname{table}{Tab.}{Tabs.}
\crefname{table}{Tab.}{Tabs.}
\Crefname{figure}{Fig.}{Figs.}
\crefname{figure}{Fig.}{Figs.}
\Crefname{equation}{Eq.}{Eqs.}
\crefname{equation}{Eq.}{Eqs.}
\Crefname{appendix}{App.}{Apps.}
\crefname{appendix}{App.}{Apps.}

\theoremstyle{plain}

\theoremstyle{definition}

\theoremstyle{remark}

\usepackage[textsize=tiny]{todonotes}

\newcounter{packednmbr}

\newenvironment{packeditemize}{\begin{list}{$\bullet$}{\setlength{\itemsep}{0.5pt}\addtolength{\labelwidth}{-4pt}\setlength{\leftmargin}{\labelwidth}\setlength{\listparindent}{\parindent}\setlength{\parsep}{1pt}\setlength{\topsep}{0pt}}}{\end{list}}

\newcommand{\calM}{\mathcal{M}}

\newcommand{\calT}{\mathcal{T}}

\icmltitlerunning{OMS-DPM: Optimizing the Model Schedule for Diffusion Probabilistic Model}

\begin{document}

\twocolumn[
\icmltitle{OMS-DPM: \textit{O}ptimizing the \textit{M}odel \textit{S}chedule for Diffusion Probabilistic Models} 



\icmlsetsymbol{equal}{*}

\begin{icmlauthorlist}
\icmlauthor{Enshu Liu}{equal,thu}
\icmlauthor{Xuefei Ning}{equal,thu}
\icmlauthor{Zinan Lin}{equal,msr}
\icmlauthor{Huazhong Yang}{thu}
\icmlauthor{Yu Wang}{thu}
\end{icmlauthorlist}

\icmlaffiliation{thu}{Department of Electronic Engineering, Tsinghua University, Beijing, China}
\icmlaffiliation{msr}{Microsoft Research, Redmond, Washinton, U.S.A}

\icmlcorrespondingauthor{Xuefei Ning}{foxdoraame@gmail.com}
\icmlcorrespondingauthor{Yu Wang}{yu-wang@mail.tsinghua.edu.cn}

\icmlkeywords{Machine Learning, ICML}

\vskip 0.3in
]



\printAffiliationsAndNotice{\icmlEqualContribution} 

\begin{abstract}
Diffusion probabilistic models (DPMs) are a new class of generative models that have achieved state-of-the-art generation quality in various domains.
Despite the promise, one major drawback of DPMs is the slow generation speed due to the large number of neural network evaluations required in the generation process.
In this paper, we reveal an overlooked dimension---model schedule---for optimizing the trade-off between generation quality and speed. 
More specifically, we observe that small models, though having worse generation quality when used alone, could outperform large models in certain generation steps.
Therefore, unlike the traditional way of using a single model, using different models in different generation steps in a carefully designed \emph{model schedule} could potentially improve generation quality and speed \emph{simultaneously}.
We design \nameshort{}, a predictor-based search algorithm, to optimize the model schedule given an arbitrary generation time budget and a set of pre-trained models.
We demonstrate that \nameshort{} can find model schedules that improve generation quality and speed than prior state-of-the-art methods across CIFAR-10, CelebA, ImageNet, and LSUN datasets.
When applied to the public checkpoints of the Stable Diffusion model, we are able to accelerate the sampling by 2$\times$ while maintaining the generation quality. 

\end{abstract}  

\section{Introduction}
\label{sec:intro}





\begin{figure}[h]
  \centering
  \includegraphics[width=0.7\linewidth]{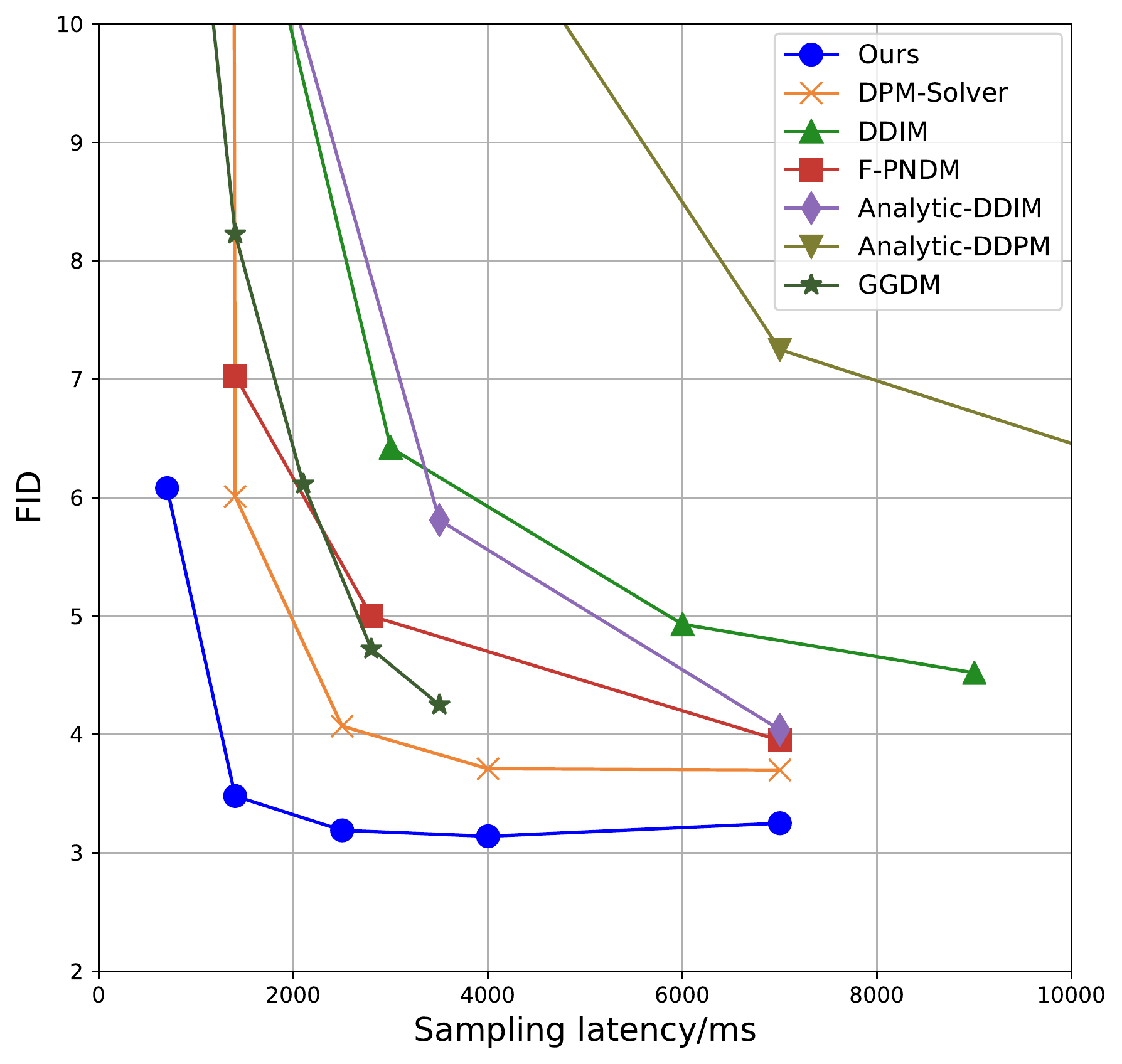}
  \caption{Generation quality v.s. latency on CIFAR-10. The horizontal axis is the time cost of generating a batch of images evaluated on a single NVIDIA A100 GPU (10 NFEs is approximately equivalent to 1400ms latency). The DPMs with model schedules derived by \nameshort{} achieves a significantly better trade-off than existing DPMs~\cite{ddim,dpm-solver,analytic-dpm,pndm,ggdm}.}
  \label{fig:comparison}
\end{figure}

Diffusion probabilistic models (DPMs) \cite{2015diffusion, ddpm, sde} are a recently emerging paradigm of generative models, which learns an iterative denoising process to transform gaussian noise to clean data. DPMs have
already outperformed \cite{diffusion_beat_gan} other alternatives like variational autoencoders (VAEs) \cite{vae} and generative adversarial networks (GANs) \cite{gan}
on both generation quality and likelihood estimation. DPMs have been successfully applied to various tasks, including image generation~\cite{ddpm,diffusion_beat_gan},
super-resolution \cite{srdiff, imagesr}, video generation \cite{videodiff}, speech generation \cite{diffwave, wavegrad}, and point cloud completion and generation \cite{diffusion3d}.

However, one major drawback of DPMs is the slow sampling speed. 
Specifically, the generation process of DPMs can be viewed as solving diffusion stochastic differential equations (SDEs) or ordinary differential equations (ODEs) using time-dependent score functions of data distributions \cite{smld, sde}. Neural networks (NNs) are trained to evaluate the score function. To solve the differential equations, DPMs usually need to discretize the continuous sample trajectories to hundreds or thousands of steps, each with one NN inference. This causes the unbearably slow sampling speed (up to 1000 times slower than GANs \cite{gan,ddim}), 
making DPMs impractical for real-time applications. 
Prior work tackles this problem mostly by proposing better denoising formation, including \emph{noise schedule}, \emph{discretization scheme}, and \emph{solver formula} ~\cite{ddim, sde, pndm, deis, dpm-solver, dp, iddpm, analytic-dpm}. 


In this paper, we point out a \emph{new} dimension for improving the trade-off between generation quality and speed---the \emph{model schedule}.
The key observation is that smaller models, though having worse generation quality when used alone, could outperform large models in certain denoising steps. Hence, unlike the common practice of using a single model, \emph{using different models for different denoising steps} could potentially lead to  benefits in \emph{both} generation quality and speed. Therefore, the \emph{model schedule}, the model assignments to each of the denoising steps, is an important factor to consider in DPMs. 

Since the training and sampling of DPMs can be decoupled \cite{ddim,sde},
using public pre-trained DPMs (e.g., Stable Diffusion \cite{ldm}) instead of training from scratch has become prevalent across academia and industry, and we expect to see more pre-trained DPMs to come in the future. Therefore, among all research directions around \emph{model schedule}, we study the following problem:
\begin{center}
    \emph{Given a set of pre-trained DPM models and a generation time budget, how can we find the model schedule that optimizes the generation quality? }
\end{center}
The problem is challenging due to the large search space that grows exponentially with respect to the number of steps.
To address the challenge, we propose a method to \underline{O}ptimize 
the \underline{M}odel \underline{S}chedule for \underline{D}iffusion \underline{P}robabilistic \underline{M}odel through predictor-based
 search (\emph{OMS-DPM}).
 Our predictor takes the model schedule as input and predicts the generation quality. The predictor is trained with a small amount of data and can generalize to unseen model schedules. Equipped with the predictor, we employ an evolutionary algorithm to quickly explore the space and derive the well-performing model schedules under a wide range of generation time budgets. 



Our contributions are as follows.

\begin{packeditemize}
\item \cref{sec:phenomenon}: We point out an overlooked dimension---model schedule---for optimizing both the generation quality and sampling speed of DPMs. Specifically, we reveal the phenomenon where globally better models do not necessarily perform better on each individual denoising step, and using different models at different steps can lead to a significant improvement in generation quality and speed. 

\item \cref{sec:method}: We propose an actionable method, \nameshort{}, to decide the model schedule that optimizes the generation quality given an arbitrary generation time budget.
As \nameshort{} focuses on a novel optimizing dimension (i.e., the model schedule), it is orthogonal and compatible with existing methods that accelerate DPM sampling, including DDIM \cite{ddim} and DPM-Solver \cite{dpm-solver}. Specifically, \nameshort{} supports searching the special parameters in these methods such as step-skipping in DDIM and the solver order in DPM-Solver.

\item \cref{sec:experiments}: We experimentally validate \nameshort{} across a wide range of datasets, including CIFAR-10, CelebA, ImageNet-64 and LSUN-Church, and show that \nameshort{} can achieve significantly better trade-offs on generation quality and speed than the baselines (\cref{fig:comparison}). For example, we are able to obtain model schedules that \emph{simultaneously} achieve better FID (3.19 v.s. 3.56) and sampling speed (2.8$\times$ times faster) than using a single model \cite{ddpm} with DPM-Solver on CIFAR-10. 

To further demonstrate the practical value, we apply \nameshort{} on the 4 public checkpoints of the popular Stable Diffusion.\footnote{https://huggingface.co/CompVis}
\nameshort{} is able to accelerate the sampling by over $2\times$ while maintaining the generation quality on text-to-image generation task on MS-COCO 256$\times$256 dataset \cite{mscoco}. We have open-sourced our code at \url{https://github.com/jsttlgdkycy/OMS-DPM} to allow the community to use \nameshort{}.

\end{packeditemize}

\section{Background and Related Work}
\label{sec:formatting}
\subsection{Diffusion Probabilistic Models}
Given a $D$-dimension random variable $x_0\in\mathbb{R}^D$, Diffusion Probabilistic Models (DPMs)~\cite{2015diffusion, ddpm} learns its distribution $q(x_0)$. DPMs define a \emph{forward diffusion process} for $x_t$~\cite{sde}: 
\begin{equation}
\mathrm{d}x_t=f(x_t, t)dt+g(t)dw_t,
\end{equation}
where $w_t$ is a standard Wiener process, and $x_0\sim q(x_0)$. $f$ and $g$ determines 
the noise magnitudes in $x_t$. The \emph{reverse diffusion process} from $T$ to $0$ is:
\begin{equation}\label{reverse_sde}
\mathrm{d}x_t=[f(t)x_t-g^2(t)\nabla_x\mathrm{log}q(x_t)]dt+g(t)\mathrm{d}\overline{w}_t,
\end{equation}
where $x_T\sim q(x_T)$ and $\overline{w}_t$ is a reverse time standard Wiener process. This SDE has an equivalent probability flow ODE:
\begin{equation}
\label{eq:probability_flow_ode}
\mathrm{d}x_t=[f(t)x_t-\frac{1}{2}g^2(t)\nabla_x\mathrm{log}q(x_t)]\mathrm{d}t,
\end{equation}
A NN (usually an U-net) is trained to learn $\nabla_x\mathrm{log}q(x_t)$. Then we can generate data by solving the reverse SDE or the ODE. \update{Most DPM methods use one single NN to evaluate the score term. While some work proposes to uses multiple NNs for function evaluation at different timesteps \cite{subspace, ediffi}, they need to train the NNs only at their corresponding timesteps. In contrast, our work can utilize existing pre-trained NNs without inducing extra training cost.}

\subsection{Training-Free Samplers} 
\label{Sec 2.3}
To solve this ODE, one should first define $f$ and $g$ (i.e., \emph{noise schedule}) and train neural networks \cite{iddpm}. We often use $\alpha_t$ and $\sigma_t$ to denote the \emph{noise schedule}, which has a relationship with $f$ and $g$ as follows:
\begin{equation}
f(t) = \frac{\mathrm{d} \mathrm{log}\alpha_t}{\mathrm{d}t}, \quad g^2(t)=\frac{\mathrm{d}\sigma^2_t}{\mathrm{d}t} - 2\frac{\mathrm{d}\mathrm{log}\alpha_t}{\mathrm{d}t}\sigma^2_t.
\end{equation}
Their ratio $\alpha_t$/$\sigma_t$ is called the signal-to-noise ratio (SNR). Then $[0,T]$ is discreted to timesteps $[t_0, t_1, \cdots, t_N]$, (i.e., \emph{discretization scheme}). Finally \emph{solver formula} is applied to compute each $x_{t_i}$ at timestep $t_i$ in order \cite{ddim,pndm,deis,dpm-solver}.
A training-free sampler only involves the last two parts of this solving process by utilzing pre-trained models.
Many training-free samplers \cite{ddim,pndm,deis,dpm-solver} have been designed to achieve better trade-offs between the number of function evaluations (NFEs) and generation quality. They can be applied to any existing network (e.g., $\epsilon_\theta(x_t, t)$) without retraining.
The following are two common samplers, both of which are compatible with our \nameshort{}. 

\textbf{DDIM} \cite{ddim} is one of the most popular samplers. Its solver formula is:
\begin{equation}\label{ddim}
\begin{aligned}
x_{t}&=\sqrt{\alpha_{t}}\frac{x_{s}-\sqrt{1-\alpha_{s}}\epsilon_\theta(x_{s}, s)}{\sqrt{\alpha_{s}}}
\\
&+\sqrt{1-\alpha_{t}-\sigma^2_{s}}\epsilon_\theta(x_{s}, s)+\sigma_{s}\epsilon_{s},
\end{aligned}
\end{equation}
where $\epsilon_{s}\sim\mathcal{N}(\epsilon|0,\textit{\textbf{I}})$. There is no restriction on the value of $s$ and $t$. However, big step sizes (small step numbers) often lead to large errors. When $\sigma_{s}$ is set to zero, this solver is called \textbf{DDIM}. 

\textbf{DPM-Solver} \cite{dpm-solver} gives a solver formula from $x_s$ to $x_t$ with a taylor expansion form :
\begin{align*}
x_t&=\frac{\alpha_t}{\alpha_s}x_s-\alpha_t\sum^{k-1}_{n=0}\epsilon^{(n)}(x_{\lambda_s},\lambda_s)\int^{\lambda_t}_{\lambda_s}e^{-\lambda}\frac{(\lambda-\lambda_s)^n}{n!}\mathrm{d}\lambda
\\
&+\mathcal{O}((\lambda_t-\lambda_s)^{k+1}).
\end{align*}
It takes $k$ NFEs to compute all derivatives. Besides, it applies a discretization scheme of uniform $\log(\mathrm{SNR})$, performing better than linear steps and quadratic steps.

\subsection{\update{AutoML}}
\update{AutoML methods aim at 
automatically deciding for the optimal machine learning system respect to specific conditions such as task, dataset, and hardware. 
The research problems in the AutoML field include model selection, hyperparameter tuning and neural architecture design \cite{oboe, pbt, rlnas}. This work mainly focus on the automatic optimization of the proposed model schedule, including model selection and sampling schedule design for DPMs.} 
\subsubsection{Predictor-based Neural Architecture Search}
Neural Architecture Search (NAS) is an important sub-task of AutoML. NAS searches for suitable architectures under certain tasks and constraints~\cite{rlnas,nas-survey}. One of the most challenging issues in NAS is the slow evaluation, due to the heavy computation burden of training and testing an architecture. Predictor-based NAS~\cite{nao,gates} is proposed to accelerate the evaluation phase, where a predictor is trained on a small number of architecture-performance pairs and can efficiently evaluate new architectures. In our problem, the evaluation of a DPM is expensive due to the slow sampling speed, and we borrow some ideas from NAS to propose a predictor-based method to accelerate the model schedule search.
\section{Model Schedule: A New Dimension in DPM Design}
\label{sec:phenomenon}

\begin{figure*}[h]
  \centering
  \begin{subfigure}{0.31\linewidth}
    \centering
    \includegraphics[width=0.9\linewidth]{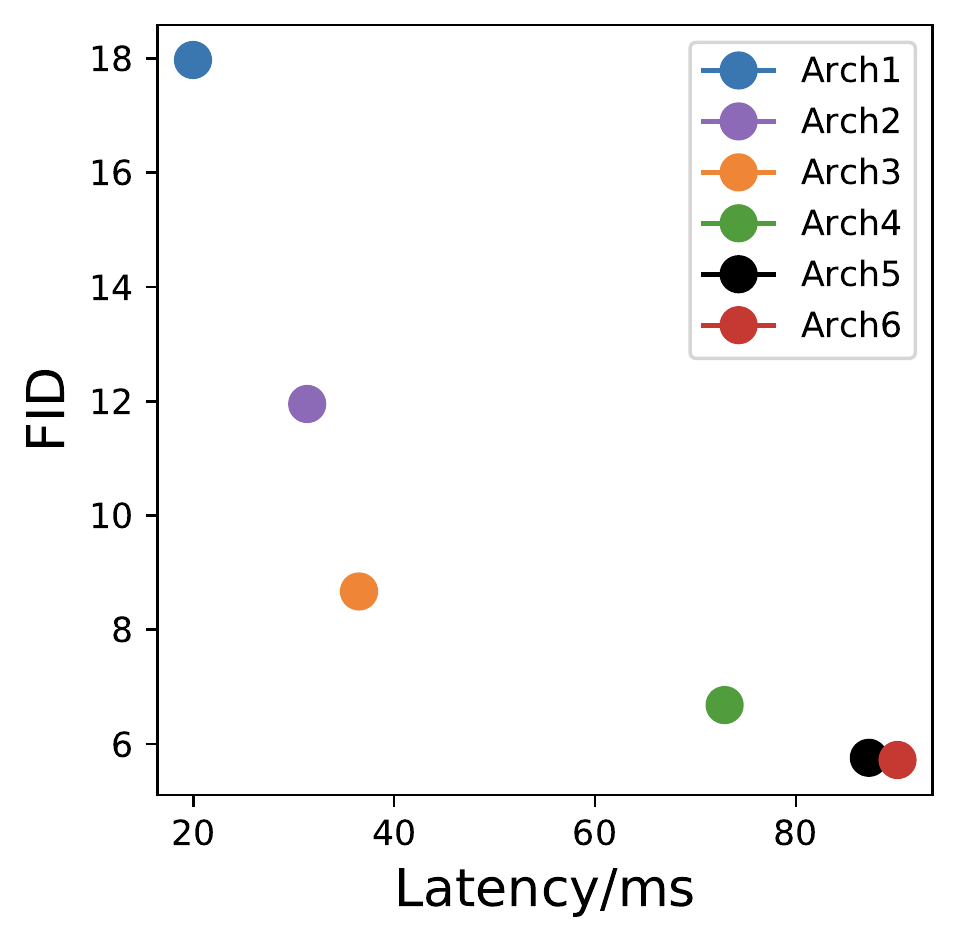}
    \caption{Basic information of model zoo. The horizontal axis stands for the latency these neural architectures take to generate a batch of 128 images, testing on a single A100 GPU. The vertical axis stands for FID evaluated on 10k images generated through a 90-step DPM-Solver sampler~\cite{dpm-solver} using these architectures alone as the score function estimator.}
    \label{fig:arch_info}
  \end{subfigure}
  \hfill
  \begin{subfigure}{0.31\linewidth}
    \centering
    \includegraphics[width=0.95\linewidth]{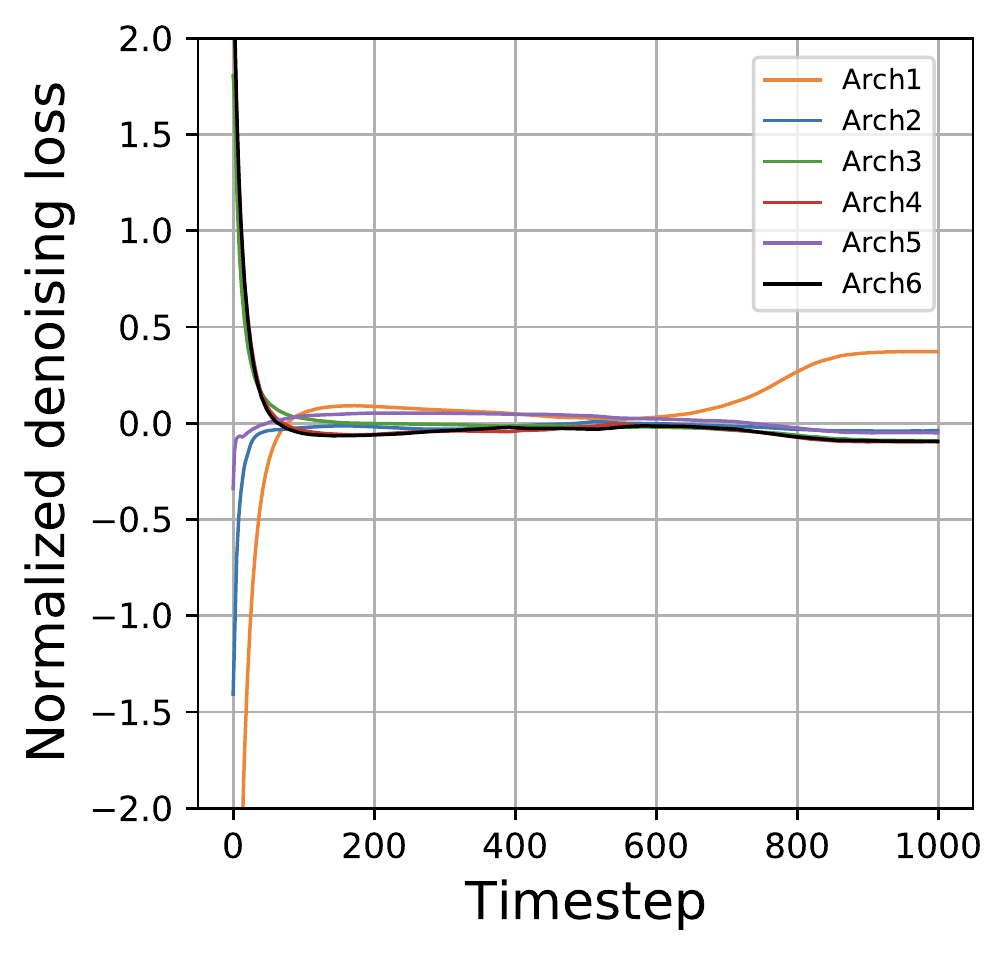}
    \caption{Denoising loss on CIFAR-10 test set. The horizontal axis stands for different steps. The vertical axis stands for the denoising loss (lower is better, normalized for each step for better visualization). The models do not have a consistent ranking across all steps. A smaller model (thus having a smaller latency) could outperform larger models in some denoising steps.}
    \label{fig:denoising_loss}
  \end{subfigure}
  \hfill
  \begin{subfigure}{0.31\linewidth}
    \centering
    \includegraphics[width=0.88\linewidth]{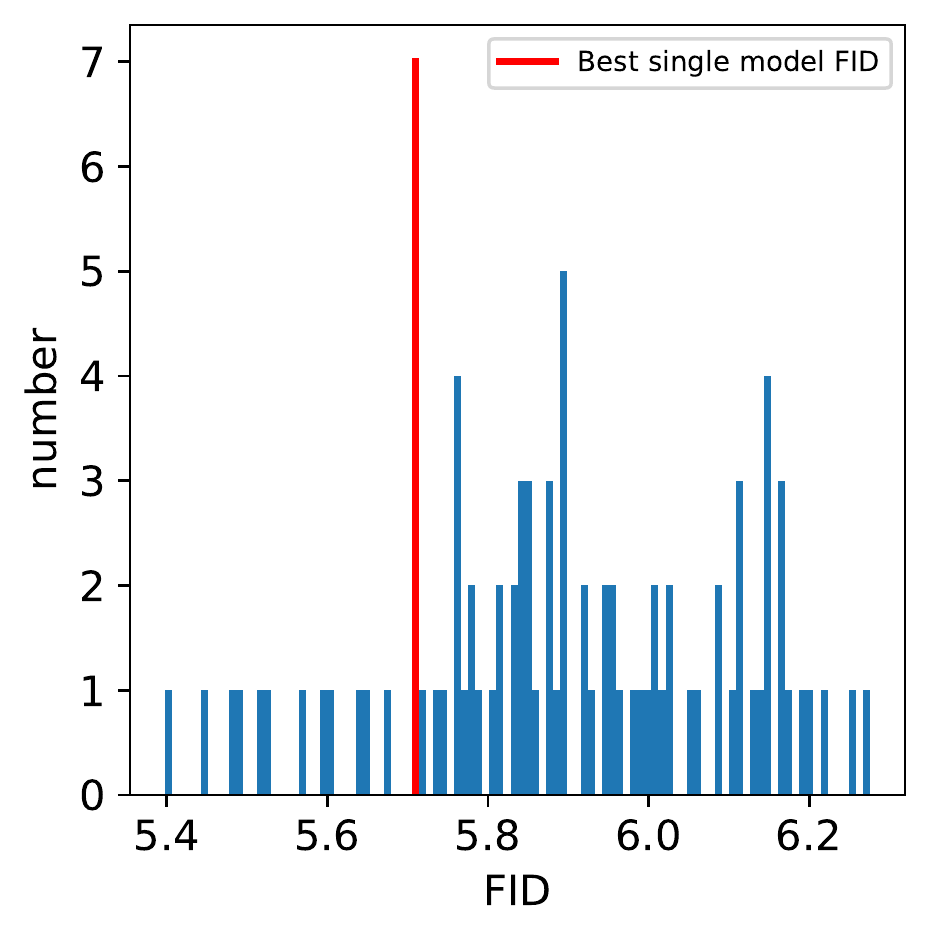}
    \caption{Histogram of FIDs obtained by randomly combining models as the score function estimators for a 90-step DPM-Solver sampler~\cite{dpm-solver}. The red line indicates the best FID achieved by a single model in the model zoo (i.e., the lowest y-axis values of all points in \cref{fig:arch_info}). This illustrates the value of mixing different models in the generation process of DPMs.}
    \label{fig:randomly_mixed}
  \end{subfigure}
  \caption{The importance of \emph{model schedule} in DPM design.}
  \label{fig:short}
\end{figure*}

Getting better trade-offs between the generation quality and speed of DPMs is an important problem that has drawn much attention. Since the only unknown term in \cref{reverse_sde} or \cref{eq:probability_flow_ode} is the score term (i.e., $\nabla_x\mathrm{log}q(x_t)$), one can modify the sampling process after getting a pre-trained model used to estimate the score. Therefore, a lot of studies~\cite{ddim,dpm-solver} have been focused on designing training-free samplers for the generation process, aiming to reduce the NFEs for generating high-quality samples.
Different from existing studies, our work reveals a new dimension---model schedule---for training-free optimization of the DPMs' generation quality and speed.
In this section, we discuss the motivation behind  \emph{model schedule}.

We start with a (perhaps surprising) observation that smaller models can actually outperform larger models at a wide range of denoising steps. 
We train a set of DPM models with different sizes on CIFAR-10 (\cref{fig:short}). Unsurprisingly, using a single network with smaller latency across all steps leads to worse sample quality due to the lower model capacity (\cref{fig:arch_info}). 
But the surprising observation is that the ranking of denoising ability is \emph{not} consistent across the whole time axis (\cref{fig:denoising_loss}). For example, architecture 1 with the worst overall generation quality actually outperforms most of the other models in the late denoising process.
\update{This observation also corroborates recent findings that different denoising steps perform different tasks \cite{slim, perception}.}

This phenomenon sheds light on a new opportunity to improve the trade-off between \emph{denoising loss} and \emph{generation latency}. For example, if we replace the late denoising process of architecture 6 with architecture 1, we will get a smaller overall (denoising) loss and a smaller generation latency \emph{simultaneously}, compared to using architecture 6 alone across all steps. 
Although a smaller denoising loss does not always indicate better generation quality (e.g., in FID) \cite{dp, soft}, we hypothesize that mixing different models across different steps could also benefit the trade-off between \emph{generation quality} and \emph{latency}.
As a proof of concept, 
we randomly pick models in each of the denoising steps (\cref{fig:randomly_mixed}). We can obtain a model (with FID $\approx 5.4$) that outperforms the model with the best generation quality in the model zoo (i.e., architecture 6) in terms of \emph{both} generation quality and speed.

The phenomenon illustrated in \cref{fig:short} opens up new interesting research questions, including understanding when and why a DPM model would favor a specific denoising step, and how we can magnify it during \emph{training} in an optimal way for the purpose of fast generation (e.g., by tuning loss weights \cite{song2021maximum}).
Given that using public pre-trained DPMs (e.g., stable diffusion \cite{ldm}) (instead of training new models from scratch) has become a popular paradigm, we study the following research question: assuming that we already have a set of pre-trained models, how we can decide the sequence of models to use, namely \emph{model schedule}, to achieve the best trade-off between generation quality and latency. 

\section{\nameshort{}: Optimizing the Model Schedule}
\label{sec:method}
\subsection{Problem Definition: Model Schedule Optimization}

Suppose we have a pre-trained model zoo $\alpha$ with $N$ models:
$\alpha=\{a_1,a_2,\cdots,a_N\}$,
where all $a_i$s are neural networks trained to predict noise~\cite{ddpm} or its variants (\cref{{sec:formatting}}). Denote the inference latency 
of model $a_i$ as $l_i$, the sampling time of the $M$-step DPM with a model schedule $a_{s_1}, a_{s_2}, \cdots, a_{s_M}$ placed on diffusion timesteps $t_1,t_2,\cdots,t_M$ can be estimated as $\sum_{m=1}^{M}l_{s_m}$, where $s_m \in \{1,\cdots,N\}$ is the model zoo index for the $m$-th denoising step.
Giving a generation time budget $C$ as the constraint, 
we aim to decide the number of sampling steps $M$ and the model schedule that optimizes the sample quality:
\begin{align}
    \mathop{\arg\min}\limits_{\substack{M\leq L,\\ a_{s_1},a_{s_2},\cdots,a_{s_M},\\ t_1,t_2,\cdots,t_M}} & \mathcal{F}([(a_{s_1},t_1),(a_{s_2},t_2),\cdots,(a_{s_M},t_M)]),\nonumber
    \\
    \text{s.t.} \ & \sum_{i=1}^{M}l_{s_i}<C
\label{eq:opt_problem}
\end{align}
where $\mathcal{F}([(a_{s_1},t_1),(a_{s_2},t_2),\cdots,(a_{s_M},t_M)])$ refers to the sample quality score (e.g., FID) of using $a_{s_1},a_{s_2},\cdots,a_{s_M}$ at timesteps $t_1,t_2,\cdots,t_M$ respectively. 
$L$ is the upper limit on the number of steps $M$.
Note that the models are applied in the order of $a_{s_M}$, $a_{s_{M-1}}$,$\cdots$,$a_{s_1}$ to transform pure noises to data, i.e., $t_1<t_2<\cdots<t_M$.

The flexible optimization space in \cref{eq:opt_problem} contains several dimensions: (1) The number of diffusion steps $M$. (2) The value of timesteps $t_i$ (i.e., the discretization scheme). (3) The model $a_{s_i}$ to apply at step $t_i$.

\begin{figure}[t]
  \centering
  \includegraphics[width=0.7\linewidth]{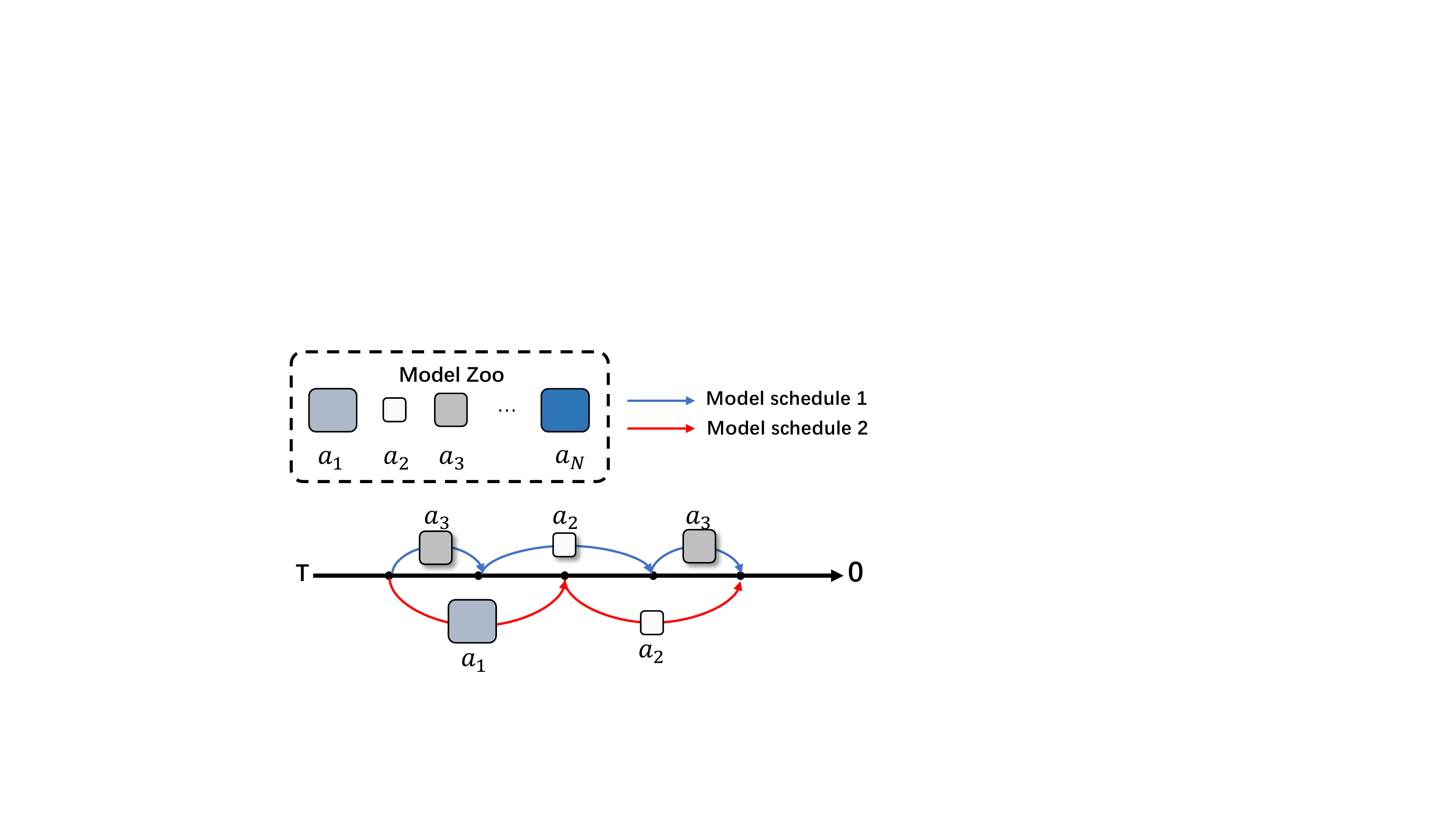}
    \vspace{-1.0em}
  \caption{Illustration of two example model schedules.}
  \label{fig:ss}
  \vspace{-1.0em}
\end{figure}

For deciding the timestep values, previous studies empirically discretize the timesteps \emph{linearly}~\cite{ddim} or following the \emph{uniform logSNR} principle~\cite{dpm-solver}. \citet{ddim} also propose a sub-sequence selection procedure that produces \emph{quadratically} discretized timesteps.
To solve \cref{eq:opt_problem}, we follow previous empirical principles to discretize the timesteps beforehand: We designate the values for all $L$ timesteps $t_1 \cdots t_L$ according to an empirical discretization scheme (\textit{linear}~\cite{ddim} for DDIM experiments, and \textit{uniform logSNR}~\cite{dpm-solver} for DPM-Solver experiments). In order to support $M\leq L$ timesteps, we introduce a special type of model into the model zoo, \emph{null model}, denoted as $a_0$. If $a_0$ is selected at a certain timestep, this timestep is unused in the reverse generation process.
In this way, our optimization space contains a vast number of flexible timestep discretizations that are different from the manually designed ones in the literature. 
For instance, \cref{fig:ss} illustrates two model schedules that can be derived from our optimization space, which have different timestep discretizations and different model choices.

After conducting the $L$-step discretization and introducing the null model, the optimization variables of problems are $\{s'_l\}_{l=1,\cdots,L}$, where $s'_l \in \{0,1,\cdots,N\}$. The size of this optimization space is $(N+1)^L$, which is extremely large, e.g., about $10^{84}$ when $L=100$ and $N=6$.

\textbf{Challenges.} One simple idea is to use the model with the smallest loss on each timestep. However, as indicated in \cite{dp}, loss values are not a good indication of the generation quality. To verify this, we run a 90-step DPM-Solver with the models with the minimal denoising loss at each step in \cref{fig:denoising_loss}. This gives an FID of 8.56, worse than the random model schedules in \cref{fig:randomly_mixed}.
A brute-force search method is also impractical due to the large search space and the evaluation overhead of $\mathcal{F}$.
For example, it takes about 1 GPU hour to generate 5k samples (256$\times$256 resolution), 
for evaluating only a \emph{single} model schedule. 

\subsection{Predictor-based Model Schedule Optimization}
To circumvent the above challenges, we propose to Optimize the Model Schedule for Diffusion Probabilistic Model through predictor-based search (OMS-DPM). 
We train a performance predictor that takes the model schedule as input and predicts its generation quality. This predictor can evaluate each model schedule in $<$1 GPU second, enabling us to solve the optimization problem efficiently.
We'll first go through the workflow of our method in \cref{sec:workflow}, and then elaborate on the predictor design in \cref{sec:predictor_design}.


\subsubsection{Overall Workflow}
\label{sec:workflow}
\begin{figure}
  \centering
  \includegraphics[width=0.97\linewidth]{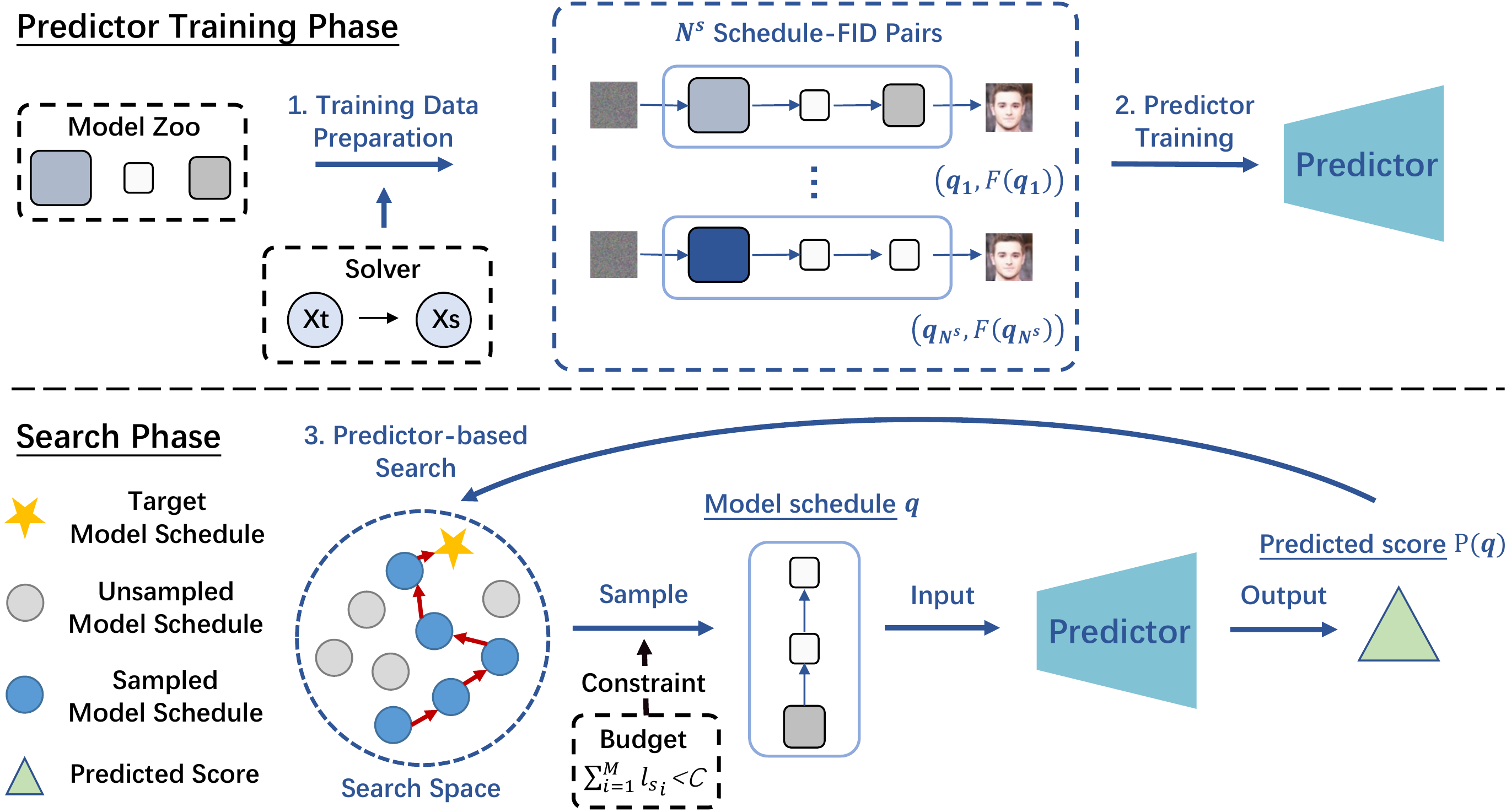}
  \caption{Our overall workflow contains 3 steps: 
  (1) Prepare the data for predictor training. (2) Train the predictor. (3) Conduct the predictor-based evolutionary search. The inputs of our workflow include \emph{model zoo}, \emph{solver type}, and \emph{budget}. 
  \emph{Budget} is only used in the search phase and does not affect the predictor training phase.} 
  \label{fig:flow}
  \hfill
\end{figure}

As shown in \cref{fig:flow}, we use the given model zoo\footnote{See \cref{Sec:model zoo} on how the model zoos can be obtained.} and ODE solver to (1) prepare the predictor training data and (2) use these data to train a performance predictor of model schedules. Then, for any given platform or budget, we can (3) run a predictor-based search to derive a suitable DPM.
In the predictor-based search, the predictor will be used to evaluate model schedules sampled by an evolutionary algorithm. We will pick the model schedule with the best predicted score while satisfying the budget constraint.


\textbf{Training Data Preparation.} 
To prepare the data for predictor training, we randomly sample $N^{s}$ 
\update{(where the superscript $s$ stands for \emph{schedule})}  model schedules $[\boldsymbol{q}_1, \cdots, \boldsymbol{q}_{N^s}]$ from the optimization space 
and evaluate their FID scores $\mathcal{F}(q_i)$. 
The sampling distribution for model schedules is designed to make the resulting sample quality scores diverse enough. To accelerate this phase, we only sample a few images to evaluate the FID. See \cref{Sec:experiment details} for the details. From this process, we get $N^s$ pairs of model schedules and FID scores $\{(\boldsymbol{q_i}, \mathcal{F}(q_i))\}_{i=1,\cdots,N^s}$.

\textbf{Predictor Training.} 
Compared with the absolute quality, the relative rankings of the model schedules are more important for guiding the search. Therefore, we adopt a pair-wise ranking loss~\cite{gates} to train the predictor:
\begin{align*}
loss=\sum^{N^s}_{i=1}\sum_{j,F(\boldsymbol{q}_j)>F(\boldsymbol{q}_i)}\max(0, m-(\mathrm{P}({\boldsymbol{q}_j)}-\mathrm{P}({\boldsymbol{q}_i}))),
\label{eq:ranking_loss}
\end{align*}
where $m$ is the hinge compare margin. 
The ranking loss drives the predictor to preserve the relative ordering of predictions $\{P(\boldsymbol{q}_i)\}$ according to the true FID scores $\{F(\boldsymbol{q}_j)\}$. When predicting for an unseen model schedule, the lower the predicted score of a model schedule $\boldsymbol{q}$ is, the better generation quality we expect $\boldsymbol{q}$ to achieve.

\textbf{Predictor-Based Evolutionary Search.} 
Thanks to the predictor, we can use any search algorithm without incurring additional DPM training or evaluation overhead. We find that a simple evolutionary algorithm~\cite{evonas} is sufficient for providing significant improvement. The algorithm iteratively modifies the current best model schedules (evaluated by the predictor) in the hope of finding better ones. 
The complete algorithm is provided in \cref{sec:app_evo}. 

Our OMS-DPM also supports searching the special parameters of the sampler (e.g., solver order in DPM-Solver); see \cref{sec:app_sampler_ss_design} for the details.

\subsubsection{Predictor Design}
\label{sec:predictor_design}

\begin{figure}
  \centering
    \centering
    \includegraphics[width=0.90\linewidth]{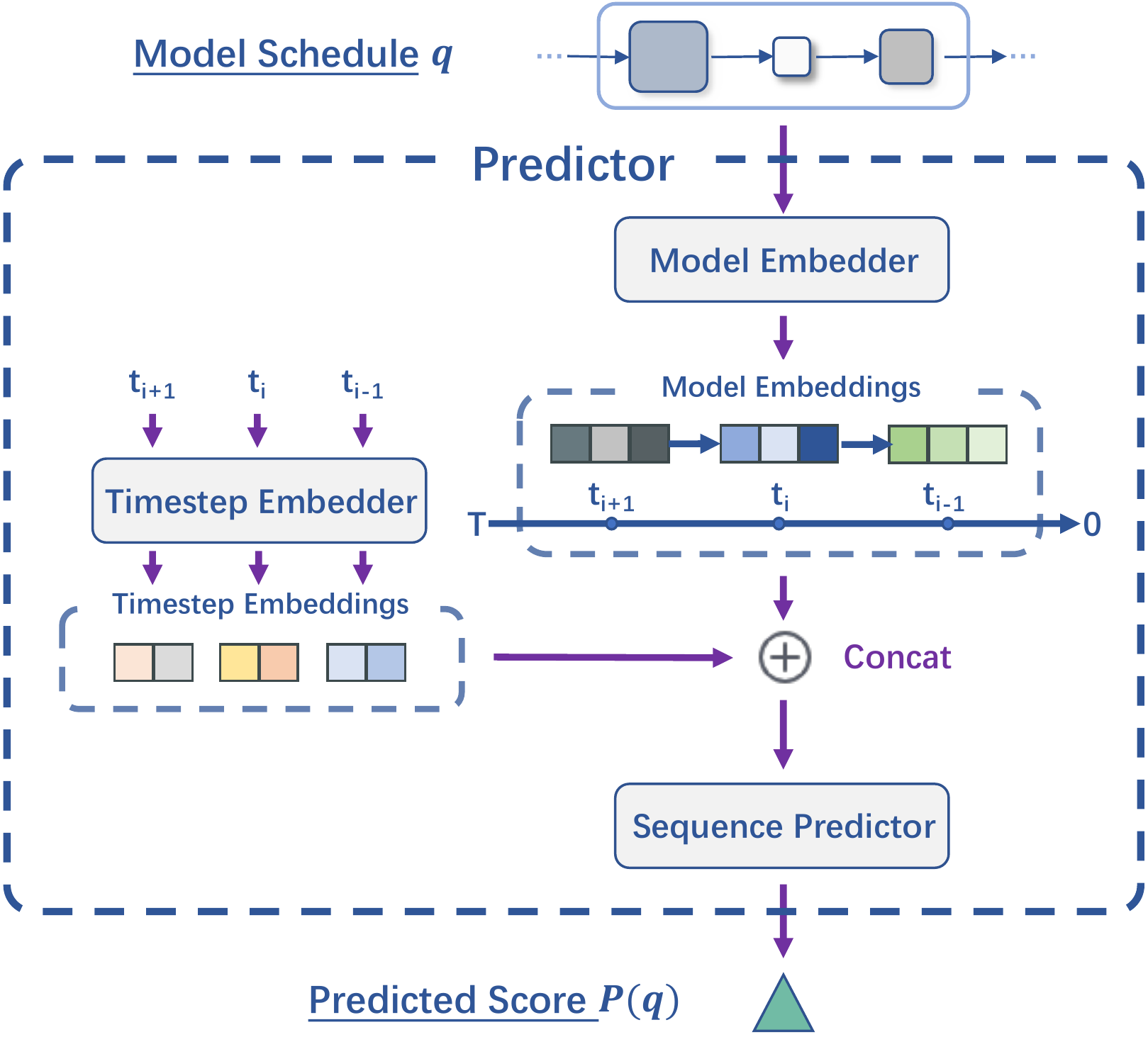}
    \vspace{-0.8em}
    \caption{
    The predictor takes a model schedule $\boldsymbol{q}$ as input and predicts the generation quality $P(\boldsymbol{q})$. The predictor consists of a model embedder, a timestep embedder, and a sequence predictor.}
    \label{fig:predictor}
  \vspace{-0.5em}
\end{figure}

Our predictor takes a model schedule $\boldsymbol{q}$ parametrized as $[s'_1,s'_2,\cdots,s'_L]$ as input and outputs a score for that schedule. As shown in \cref{fig:predictor}, the predictor consists of a model embedder, a timestep embedder, and a sequential predictor.

\textbf{Model Embedder.} The model embedder maps a model choice $s'_l \in \{0,1,\cdots,N\}$ to a $d^\calM$-dimension continuous embedding $\mathrm{Emb}^\calM_l \in \mathbb{R}^{d^\calM}$ ($\calM$ stands for \emph{model}). Each model 
corresponds to a row in the globally trainable embedding matrix $\boldsymbol{O}\in\mathbb{R}^{(N+1) \times d^\calM}$. 

\textbf{Timestep Embedder.} 
The denoising functionality at each step is not only related to the model choice, but also the current timestep. Therefore, besides the model embedding, it is a logical choice to add timestep embedding as an additional input to our predictor. 
Our timestep encoder is an MLP that gets the sinusoidal embedding vector~\cite{transformer,ddpm} from a single timestep scalar $t_l$ and outputs the embedding $\mathrm{Emb}^\calT_{l}\in\mathcal{R}^{d^\calT}$ ($\calT$ stands for timestep). 

\textbf{Sequence Predictor.} After getting the model embeddings $[\mathrm{Emb}^\calM_1, \cdots, \mathrm{Emb}^\calM_L]$ and the timestep embeddings $[\mathrm{Emb}^\calT_1, \cdots, \mathrm{Emb}^\calT_L]$, 
we concatenate them at each timestep $l$ to get $\mathrm{Feat}_l = (\mathrm{Emb}^\calM_l|\; \mathrm{Emb}^\calT_l)$. Then the sequence $[\mathrm{Feat}_1, \cdots, \mathrm{Feat}_L]$ is input into an LSTM. The output features of the LSTM are averaged across timesteps and fed into an MLP to get the final score $\mathrm{P}(\boldsymbol{q})$.



\section{Experiments}
\label{sec:experiments}
In this section, we first show the 
results
of \nameshort{} with two popular samplers, DPM-Solver~\cite{dpm-solver} and DDIM~\cite{ddim}, on CIFAR-10, CelebA, ImageNet-64, and LSUN-Church datasets (\cref{ddim_dpm_experiments}). 
We then demonstrate the practical value of \nameshort{} with Stable Diffusion (\cref{sec:stable-diffusion}). Ablation studies are provided in \cref{sec:ablation}.
We provide our empirical insights in \cref{sec:empirical_obs}. 
The model zoo information and experimental settings 
can be found in \cref{Sec:model zoo} and \cref{Sec:experiment details}. 

\begin{table*}[t]
  \centering
  \begin{subtable}[t]{0.495\linewidth}
    \centering
    \begin{tabular}{ccccc}
      \toprule
      \multirow{2}{*}{\textbf{Budget/ms}} & \multicolumn{3}{c}{\textbf{Baseline Type}} & \multirow{2}{*}{\textbf{Ours}}\\
      \cmidrule(lr){2-4} 
       & \textbf{(1)} & \textbf{(2)} & \textbf{(3)} & \\
      \midrule
      $7.0\times10^3$ & 3.56 & 3.33 & 9.33$\pm$0.17 & \textbf{3.25$\pm$0.01}\\
      $4.0\times10^3$ & 3.61 & 3.33 & 11.77$\pm$0.53 & \textbf{3.14$\pm$0.02}\\
      $2.5\times10^3$ & 3.93 & 3.64 & 13.61$\pm$0.37 & \textbf{3.19$\pm$0.05}\\
      $1.4\times10^3$ & 5.23 & 6.40 & 17.42$\pm$2.09 & \textbf{3.48$\pm$0.06}\\
      $0.7\times10^3$ & 8.73 & 10.68 & 24.11$\pm$3.91 & \textbf{6.08$\pm$0.00}\\
      \bottomrule
    \end{tabular}
    \label{tab:DPM-solver+CIFAR-10}
    \caption{Results on CIFAR-10}
  \end{subtable}
  \begin{subtable}[t]{0.495\linewidth}
    \begin{tabular}{ccccc}
      \toprule
      \multirow{2}{*}{\textbf{Budget/ms}} & \multicolumn{3}{c}{\textbf{Baseline Type}} & \multirow{2}{*}{\textbf{Ours}}\\
      \cmidrule(lr){2-4} 
       & \textbf{(1)} & \textbf{(2)} & \textbf{(3)} & \\
      \midrule
      $7.0\times10^3$ & 2.49 & 2.41 & 2.89$\pm$0.06 & \textbf{2.13$\pm$0.03}\\
      $5.0\times10^3$ & 2.49 & 2.79 & 3.64$\pm$0.29 & \textbf{2.12$\pm$0.03}\\
      $3.0\times10^3$ & 2.40 & 2.76 & 6.27$\pm$0.64 & \textbf{2.17$\pm$0.03}\\
      $1.5\times10^3$ & 2.78 & 3.07 & 7.75$\pm$0.94 & \textbf{2.42$\pm$0.06}\\
      $0.65\times10^3$ & 4.79 & 6.19 & 11.27$\pm$2.08 & \textbf{3.53$\pm$0.22}\\
      \bottomrule
    \end{tabular}
    \label{tab:DPM-solver+celeba}
    \caption{Results on CelebA}
  \end{subtable}
  
  \begin{subtable}[t]{0.5\linewidth}
    \begin{tabular}{ccccc}
      \toprule
      \multirow{2}{*}{\textbf{Budget/ms}} & \multicolumn{3}{c}{\textbf{Baseline Type}} & \multirow{2}{*}{\textbf{Ours}}\\
      \cmidrule(lr){2-4} 
       & \textbf{(1)} & \textbf{(2)} & \textbf{(3)} & \\
      \midrule
      $12.0\times10^3$ & 12.99 & 13.03 & 19.88$\pm$1.62 & \textbf{12.86$\pm$0.08}\\
      $8.0\times10^3$ & 13.44 & 13.38 & 25.77$\pm$2.21 & \textbf{13.01$\pm$0.10}\\
      $5.0\times10^3$ & 14.00 & 14.33 & 30.79$\pm$1.44 & \textbf{13.64$\pm$0.13}\\
      $2.0\times10^3$ & 18.20 & 19.25 & 37.70$\pm$2.58 & \textbf{16.77$\pm$0.31}\\
      $0.8\times10^3$ & 29.59 & 43.21 & 47.62$\pm$2.13 & \textbf{23.94$\pm$0.00}\\
      \bottomrule
    \end{tabular}
    \label{tab:DPM-solver+imagenet64}
    \caption{Results on ImageNet-64}
  \end{subtable}
  \begin{subtable}[t]{0.495\linewidth}
    \begin{tabular}{ccccc}
      \toprule
      \multirow{2}{*}{\textbf{Budget/ms}} & \multicolumn{3}{c}{\textbf{Baseline Type}} & \multirow{2}{*}{\textbf{Ours}}\\
      \cmidrule(lr){2-4} 
       & \textbf{(1)} & \textbf{(2)} & \textbf{(3)} & \\
      \midrule
      $35\times10^3$ & 11.97 & 10.79 & 22.53$\pm$0.84 & \textbf{9.30$\pm$0.01}\\
      $25\times10^3$ & 12.02 & 10.79 & 40.55$\pm$7.02 & \textbf{9.30$\pm$0.01}\\
      $15\times10^3$ & 12.03 & 10.84 & 47.87$\pm$14.31 & \textbf{9.39$\pm$0.11}\\
      $10\times10^3$ & 13.23 & 10.84 & 68.53$\pm$14.28 & \textbf{9.25$\pm$0.00}\\
      $4\times10^3$ & 32.23 & 32.57 & 135.05$\pm$11.59 & \textbf{13.94$\pm$0.00}\\
      \bottomrule
    \end{tabular}
    \label{tab:DPM-solver+church}
    \caption{Results on LSUN-Church}
  \end{subtable}
  \caption{FIDs of our searched schedules on four datasets with DPM-Solver. Budget stands for the time cost limit of generating a batch of images. We report our results against three baselines mentioned before: \textbf{(1)} Using a single model in the model zoo and changing the NFE to meet the budget constraint. \textbf{(2)} The best schedule in the training set of the predictor. \textbf{(3)} Random sampling from the search space.}
  \label{tab:DPM-solver}
\end{table*}

\begin{table*}[t]
  \centering
  \begin{subtable}[t]{0.495\linewidth}
    \begin{tabular}{ccccc}
      \toprule
      \multirow{2}{*}{\textbf{Budget/ms}} & \multicolumn{3}{c}{\textbf{Baseline Type}} & \multirow{2}{*}{\textbf{Ours}}\\
      \cmidrule(lr){2-4} 
       & \textbf{(1)} & \textbf{(2)} & \textbf{(3)} & \\
      \midrule
      $9.0\times10^3$ & 4.29 & 4.19 & 7.67$\pm$0.26 & \textbf{3.80$\pm$0.06}\\
      $6.0\times10^3$ & 4.73 & 4.54 & 8.86$\pm$0.78 & \textbf{4.07$\pm$0.06}\\
      $3.0\times10^3$ & 6.42 & 7.10 & 13.14$\pm$1.19 & \textbf{5.20$\pm$0.01}\\
      $1.5\times10^3$ & 10.01 & 9.72 & 15.55$\pm$1.02 & \textbf{8.24$\pm$0.20}\\
      $0.75\times10^3$ & 16.11 & 14.75 & 22.04$\pm$6.07 & \textbf{12.34$\pm$0.31}\\
      \bottomrule
    \end{tabular}
    \label{tab:ddim+CIFAR-10}
    \caption{Results on CIFAR-10}
  \end{subtable}
  \begin{subtable}[t]{0.495\linewidth}
    \begin{tabular}{ccccc}
      \toprule
      \multirow{2}{*}{\textbf{Budget/ms}} & \multicolumn{3}{c}{\textbf{Baseline Type}} & \multirow{2}{*}{\textbf{Ours}}\\
      \cmidrule(lr){2-4} 
       & \textbf{(1)} & \textbf{(2)} & \textbf{(3)} & \\
      \midrule
      $15\times10^3$ & 4.61 & 4.51 & 5.97$\pm$1.32 & \textbf{3.62$\pm$0.06}\\
      $10\times10^3$ & 4.75 & 4.73 & 7.49$\pm$0.86 & \textbf{3.71$\pm$0.04}\\
      $7.0\times10^3$ & 5.03 & 5.75 & 8.21$\pm$0.64 & \textbf{3.99$\pm$0.06}\\
      $4.0\times10^3$ & 5.64 & 7.01 & 10.34$\pm$0.77 & \textbf{4.75$\pm$0.03}\\
      $1.5\times10^3$ & 7.32 & 10.31 & 12.21$\pm$2.22 & \textbf{7.07$\pm$0.13}\\
      \bottomrule
    \end{tabular}
    \label{tab:ddim+celeba}
    \caption{Results on CelebA}
  \end{subtable}
  
  \begin{subtable}[t]{0.495\linewidth}
    \begin{tabular}{ccccc}
      \toprule
      \multirow{2}{*}{\textbf{Budget/ms}} & \multicolumn{3}{c}{\textbf{Baseline Type}} & \multirow{2}{*}{\textbf{Ours}}\\
      \cmidrule(lr){2-4} 
       & \textbf{(1)} & \textbf{(2)} & \textbf{(3)} & \\
      \midrule
      $20\times10^3$ & 14.74 & 14.78 & 20.08$\pm$0.63 & \textbf{14.67$\pm$0.07}\\
      $15\times10^3$ & 15.12 & 14.81 & 23.84$\pm$0.65 & \textbf{14.78$\pm$0.08}\\
      $10\times10^3$ & 16.12 & 16.42 & 26.77$\pm$2.01 & \textbf{15.16$\pm$0.10}\\
      $5\times10^3$ & 19.47 & 20.57 & 33.11$\pm$2.51 & \textbf{18.07$\pm$0.16}\\
      $2\times10^3$ & 26.91 & 30.48 & 40.44$\pm$0.87 & \textbf{25.10$\pm$0.52}\\
      \bottomrule
    \end{tabular}
    \label{tab:ddim+iamgenet64}
    \caption{Results on ImageNet-64}
  \end{subtable}
  \begin{subtable}[t]{0.495\linewidth}
    \begin{tabular}{ccccc}
      \toprule
      \multirow{2}{*}{\textbf{Budget/ms}} & \multicolumn{3}{c}{\textbf{Baseline Type}} & \multirow{2}{*}{\textbf{Ours}}\\
      \cmidrule(lr){2-4} 
       & \textbf{(1)} & \textbf{(2)} & \textbf{(3)} & \\
      \midrule
      $55\times10^3$ & 11.72 & 11.24 & 23.33$\pm$1.04 & \textbf{10.95$\pm$0.08}\\
      $40\times10^3$ & 11.73 & 11.65 & 31.22$\pm$0.70 & \textbf{10.98$\pm$0.11}\\
      $25\times10^3$ & 12.05 & 13.15 & 40.59$\pm$9.16 & \textbf{11.10$\pm$0.22}\\
      $10\times10^3$ & 14.77 & 17.80 & 50.28$\pm$13.17 & \textbf{13.70$\pm$0.20}\\
      $4\times10^3$ & 25.27 & 67.49 & 57.97$\pm$4.55 & \textbf{23.03$\pm$0.67}\\
      \bottomrule
    \end{tabular}
    \label{tab:ddim+church}
    \caption{Results on LSUN-Church}
  \end{subtable}
  \caption{FIDs of our searched schedules on four datasets with DDIM.}
  \label{tab:ddim}
\end{table*}

\textbf{Baselines.} We use three types of baselines as listed below.
 \textbf{Baseline (1)} \update{uses a single model across all timesteps and adopts common sampler settings (including timestep schedules or DPM-solver orders)}. Specifically, we first compute the steps $S$ of using model $a_i$ according to $S(C,i) = ceil(C/l_i)$ at each constraint $C$, where $l_i$ means the latency of $a_i$ and $ceil$ means upper rounding. We report the best FID among all models and in the model zoo \update{using all sampler settings } at every budget constraint $C$. Complete results \update{and implementation details} can be found at \cref{Sec:details baseline1}.
 \textbf{Baseline (2)} 
 is the \update{model schedule} with the best FID 
 under budget constraint $C$ in the training set. 
 \textbf{Baseline (3)} is a randomly generated model schedule under budget constraint $C$. 
 We run this process for 3 random seeds.
The comparison with baseline \textbf{(1)} can illustrate the 
importance 
of mixing models, while the comparison with baseline \textbf{(2)} and \textbf{(3)} can illustrate the necessity of using a predictor and a search algorithm. 

For our \nameshort{}, we run the search process with 3 random seeds and report the mean FID and the standard deviation. Information about the predictor can be found in \cref{tab:predictor}. All predictors achieve a high Kendall's Tau (KD) with unseen validation data, indicating their reliability. 





\subsection{The Effectiveness of \nameshort{}}
\label{ddim_dpm_experiments}
\cref{tab:DPM-solver,tab:ddim} show the results with DPM-Solver and DDIM respectively.
The key takeaways are:

\textbf{The importance of model schedules.}
We can see that baseline (2) outperforms baseline (1) in many cases. This further confirms the observation in \cref{fig:randomly_mixed} where mixing multiple models in the generation process is better than the current practice of using a single model.

\textbf{The importance of the predictor and the search algorithm in \nameshort{}.} We see that our \nameshort{} always outperforms baseline (2) and baseline (3) by a large margin. These benefits indicate that our predictor is able to generalize from the limited training set, and the search algorithm is able to find a better model schedule that outperforms the best one in the training set.

\textbf{The robustness of \nameshort{} across datasets, budgets, and samplers.}
Our \nameshort{} always outperforms baseline (1) across all datasets, budgets, and both samplers. For example, 
under DPM-Solver, \nameshort{} achieves a significant boost under low budgets (e.g., 6.08 v.s. 8.73 on CIFAR-10, 23.94 v.s. 29.59 on ImageNet-64, 13.94 v.s. 32.23 on LSUN-Church). As the budget increases and more NFEs can be used for the generation, the FID will decrease and converge to a value. We can see that \nameshort{} can further lower the converging FID of DPM-Solver by properly mixing models (e.g., 3.14 v.s. 3.56 on CIFAR-10,  9.25 v.s. 11.97 on LSUN-Church).
\cref{fig:comparison} provides a comparison to more state-of-the-art methods, where we report the results of DDIM and DPM-Solver with our own implementation and take other results from the original paper~\cite{analytic-dpm,pndm,ggdm}.
We see that \nameshort{} clearly achieves the best tradeoffs between generation quality and speed. 
These results demonstrate that \nameshort{} is robust across datasets and budgets, and works well with state-of-the-art samplers.


\subsection{Results with Stable Diffusion}
\label{sec:stable-diffusion}
To further demonstrate the practical value of \nameshort{}, 
we test \nameshort{} on the text-to-image generation task using Stable Diffusion \cite{ldm}. We use the four officially released models to construct the model zoo, and choose DPM-Solver as the sampler. We test our FID on MS-COCO 256$\times$256 validation set \cite{mscoco}. 
Because the four models share the same architecture and thus have the same latency, the time cost only depends on the number of timesteps, and we show NFE as the time cost budget. Detailed settings can be found at \cref{Sec:experiment details}.

\textbf{Practical value of \nameshort{}.} Tab.~\ref{tab:sd} shows that \nameshort{} can achieve a better FID with 12 steps than the best single model with 24 steps. Given the popularity of Stable Diffusion, \nameshort{} can potentially save a significant amount of computation resources for the community. 

\textbf{More insights on model schedule and more use cases of \nameshort{}.} Note that the 4 checkpoints used here are from different training iterations of \emph{the same model}. Therefore, this experiment generalizes our insights in \cref{sec:phenomenon}: 
the phenomenon in \cref{fig:denoising_loss} happens for \update{different models with the same neural architecture}, and mixing checkpoints from \update{a single training process} with \nameshort{} is also beneficial. This insight could lead to broader use cases of \nameshort{}: since it is common to save multiple checkpoints during the training process, any developer or user of DPMs can use \nameshort{} to boost the generation speed and quality.
\begin{table}[t]
  \centering
  \begin{tabular}{c|ccc}
    \toprule
    \multirow{2}{*}{\textbf{Budget/NFE}} & \multicolumn{2}{c}{\textbf{Baseline Type}} & \multirow{2}{*}{\textbf{Ours}}\\
    \cmidrule(lr){2-3} 
    & \textbf{(1)} & \textbf{(2)} & \\
    \midrule
    9 & 13.01 & 13.01 & \textbf{12.90}\\
    12 & 12.11 & 11.37 & \textbf{11.34}\\
    15 & 11.92 & 11.13 & \textbf{10.72} \\
    18 & 11.88 & 11.13 & \textbf{10.68} \\
    24 & 11.81 & 11.13 & \textbf{10.57}\\
    \bottomrule
  \end{tabular}
  \caption{FID on MS-COCO 256$\times$256. All FIDs in the table are calculated between 30k images in validation set and 30k generated images guided with the same captions.}
  \label{tab:sd}
\end{table}

\subsection{Ablation Study}
\label{sec:ablation}
We study the influences of \emph{model zoo size} and the \emph{predictor training data size} using DPM-Solver on CIFAR-10.

\textbf{Model Zoo Size $N$.}
\cref{tab:ablation model zoo} shows the FID results of \nameshort{} with different model zoo sizes. We can see that using the largest model zoo with $N=6$ achieves the best results, and we see an improvement over the baseline across all model zoo sizes. Nevertheless, the $N=2$ results are slightly better than $N=4$. One potential reason is that $N=2$ induces a smaller search space, and thus \nameshort{} can explore the search space more sufficiently.

\begin{table}[t]
  \centering
  \resizebox{0.45\textwidth}{!}{
  \begin{tabular}{c|c|ccc}
       \toprule
       \multirow{2}{*}{\textbf{Budget/ms}} & \multirow{2}{*}{Manner} & \multicolumn{3}{c}{Model Zoo Size} \\
       \cmidrule(lr){3-5}
        & & \textbf{2} & \textbf{4} & \textbf{6} \\
       \midrule
       \multirow{2}{*}{$7.0\times10^3$} & Baseline (2) & 3.44 & 3.68 & 3.33 \\
       \cmidrule(lr){2-2}
       & Search & \textbf{3.37$\pm$0.01} & \textbf{3.49$\pm$0.02} & \textbf{3.25$\pm$0.01} \\
       \midrule
       \multirow{2}{*}{$4.0\times10^3$} & Baseline (2) & 3.57 & 3.59 & 3.33 \\
       \cmidrule(lr){2-2}
       & Search & \textbf{3.36$\pm$0.02} & \textbf{3.42$\pm$0.00} & \textbf{3.14$\pm$0.02} \\
       \midrule
       \multirow{2}{*}{$2.5\times10^3$} & Baseline (2) & 3.63 & 3.59 & 3.64 \\
       \cmidrule(lr){2-2}
       & Search & \textbf{3.29$\pm$0.01} & \textbf{3.39$\pm$0.04} & \textbf{3.19$\pm$0.05} \\
       \midrule
       \multirow{2}{*}{$1.4\times10^3$} & Baseline (2) & 6.01 & 6.58 & 6.40 \\
       \cmidrule(lr){2-2}
       & Search & \textbf{3.99$\pm$0.01} & \textbf{3.67$\pm$0.03} & \textbf{3.48$\pm$0.06} \\
       \midrule
       \multirow{2}{*}{$0.7\times10^3$} & Baseline (2) & 28.15 & 13.88 & 10.68 \\
       \cmidrule(lr){2-2}
       & Search & \textbf{6.05$\pm$0.00} & \textbf{6.25$\pm$0.36} & \textbf{6.08$\pm$0.00} \\
       \bottomrule
  \end{tabular}
  }
  \caption{Searched FIDs of composing fewer models.}
  \label{tab:ablation model zoo}
\end{table}
\begin{figure*}[h]
  \centering
  \begin{subfigure}{0.8\linewidth}
    \centering
    \includegraphics[width=0.8\linewidth]{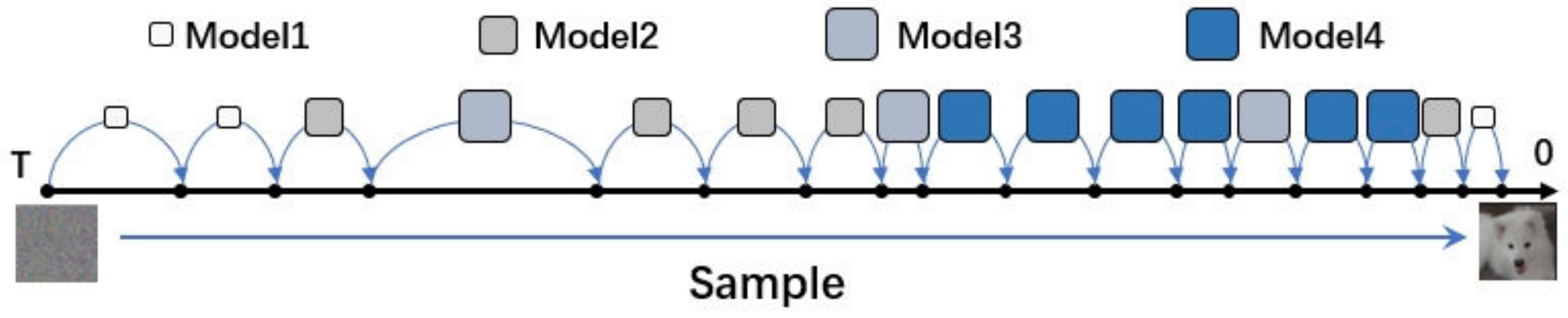}
    \caption{A searched schedule on ImageNet-64 using DDIM under 2000ms budget.}
  \end{subfigure}
  \vfill
  \begin{subfigure}{0.8\linewidth}
    \centering
    \includegraphics[width=0.8\linewidth]{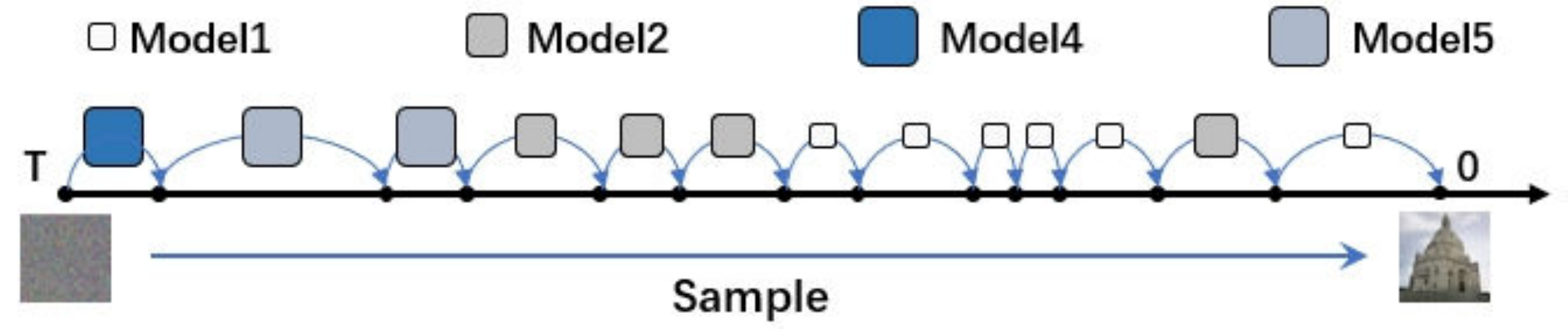}
    \caption{A searched schedule on LSUN-Church using DDIM under 4000ms budget.}
  \end{subfigure}
  \caption{Two examples of searched schedules. The model numbers shown in the figure are consistent with the description at \cref{Sec:model zoo}. The sizes of model squares match the latencies of the corresponding models approximately.}
  \label{fig:search_pattern}
\end{figure*}
\textbf{Predictor Training Data Size.} \cref{tab:ablation training data} shows the FID results using different data sizes for predictor training. Unsurprisingly, using less data to train the performance predictor results in degraded performances.
But in all three cases, \nameshort{} still achieves better results than the baseline, indicating that we can make the predictor training phase more efficient while still obtaining promising results.

\subsection{Empirical Observations}
\label{sec:empirical_obs}
We summarize the observations on the model schedule pattern, hoping to provide some practical insights. \update{Two of our searched schedules are demonstrated in \cref{fig:search_pattern}}. More searched patterns can be found in \cref{fig:search_pattern_full}. 

\textbf{Should we use more steps or larger models?} Under a low time budget, using smaller models with more steps is more likely to gain a better generation quality than using larger models with fewer steps.
The model schedules derived by \nameshort{} under the tightest budget (the last row of \cref{tab:DPM-solver} and \cref{tab:ddim}) are composed of only the 2 or 3 models with the lowest latency in the model zoo. This is because when the total number of steps is small, the error caused by the inexact solver formula and time discretization rises very quickly as the NFE decreases. Conversely, when having an adequate time budget, using larger models is suggested. 

\textbf{Should we apply larger models earlier or later?} On ImageNet-64 with DPM-Solver and DDIM samplers and on CIFAR-10 with the DDIM sampler, using large models at steps near the generated data is more likely to achieve better performances than using large models at steps near the noise. Nevertheless, on LSUN-Church, the case is just the opposite. That is, using large models at steps near the noise is more likely to achieve better performances. 

\textbf{How should the timestep be discretized?} For DDIM, we find the step size of our discovered schedules is larger at steps near the final generated image on CIFAR-10, CelebA, and ImageNet-64, especially on CelebA. On LSUN-Church, the step sizes at both ends are usually smaller than those in the middle part.
 
\textbf{Which solver order should be used?} For DPM-Solver, most of the discovered schedules apply the 1-st or 2-nd solvers under tight budgets. The 3-rd solver is only used under an adequate budget. For Stable Diffusion, our discovered schedules are mixed with 1-st, 2-nd, and 3-rd solvers under all budgets, and the 1-st solver is preferred when $t$ is close to 0.

\section{Limitations \update{and Future Work}}
Although our method can efficiently derive specialized DPMs for any given budget, if given a new dataset or task, we need to prepare the predictor training data on that dataset or task and train a new predictor, which incurs a substantial overhead. Extending our method to be capable of efficiently deriving DPMs for new datasets and downstream tasks is an interesting future direction. Besides, as our experiments in \cref{tab:ablation model zoo} demonstrate that the quality and size of the model zoos matter for the performances, how to efficiently construct a good model zoo is a topic worth studying. \update{For example, can we efficiently prune a pretrained model to get a diverse model zoo? Or can we design the ELBO loss weight \cite{perception, ELBOweights} to train a diverse model zoo?}

\update{
Finally, let us take a broader perspective than deciding the best model schedule of pretrained models in DPM: 
As more and more open-source or proprietary models and APIs with varying expertise and complexity are coming forth, we believe the idea of cleverly combining off-the-shelf models and APIs to improve performance-efficiency trade-offs can support a wider range of applications.
}

\section*{Acknowledgements}
This work was supported by National Natural Science Foundation of China (No. U19B2019, 61832007), Tsinghua University Initiative Scientific Research Program, Beijing National Research Center for Information Science and Technology (BNRist), Tsinghua EE Xilinx AI Research Fund, and Beijing Innovation Center for Future Chips. We thank Prof. Jianfei Chen, Tianchen Zhao, Junbo Zhao for their discussion.

\clearpage
\bibliography{example_paper}
\bibliographystyle{icml2023}

\newpage
\appendix
\onecolumn
\section{Additional Results}
\subsection{Results of ablation study}
The results of our ablation study with less predictor training data are shown in \cref{tab:ablation training data}.
\begin{table}[t]
  \centering
  \resizebox{0.45\textwidth}{!}{
  \begin{tabular}{c|c|cccc}
       \toprule
       \multirow{2}{*}{\textbf{Budget/ms}} & \multirow{2}{*}{Manner} & \multicolumn{3}{c}{Training Data} \\
       \cmidrule(lr){3-5}
        & & \textbf{915} & \textbf{1831} & \textbf{3662} \\
       \midrule
       \multirow{2}{*}{$7.0\times10^3$} & Baseline (2) & 3.49 & \textbf{3.33} & 3.33 \\
       \cmidrule(lr){2-2}
       & Search & \textbf{3.46$\pm$0.06} & 3.34$\pm$0.07 & \textbf{3.25$\pm$0.01}\\
       \midrule
       \multirow{2}{*}{$4.0\times10^3$} & Baseline (2) & 3.43 & 3.33 & 3.33 \\
       \cmidrule(lr){2-2}
       & Search &  \textbf{3.39$\pm$0.01} & \textbf{3.28$\pm$0.05} & \textbf{3.14$\pm$0.02}\\
       \midrule
       \multirow{2}{*}{$2.5\times10^3$} & Baseline (2) & 3.64 & 3.64 & 3.64 \\
       \cmidrule(lr){2-2}
       & Search & \textbf{3.51$\pm$0.05} & \textbf{3.41$\pm$0.11} & \textbf{3.19$\pm$0.05}\\
       \midrule
       \multirow{2}{*}{$1.4\times10^3$} & Baseline (2) & 7.41 & 7.01 & 6.40 \\
       \cmidrule(lr){2-2}
       & Search &  \textbf{3.97$\pm$0.02} & \textbf{3.75$\pm$0.14}  & \textbf{3.48$\pm$0.06}\\
       \midrule
       \multirow{2}{*}{$0.7\times10^3$} & Baseline (2) & 10.68 & 10.68 & 10.68\\
       \cmidrule(lr){2-2}
       & Search &  \textbf{6.48$\pm$0.46} & \textbf{7.53$\pm$0.07} & \textbf{6.08$\pm$0.00} \\
       \bottomrule
  \end{tabular}
  }
  \caption{Searched FIDs of using less data for predictor training.}
  \label{tab:ablation training data}
\end{table}

\subsection{Demonstration of searched model schedules}
\begin{figure*}[h]
  \centering
  \begin{subfigure}{0.8\linewidth}
    \centering
    \includegraphics[width=0.8\linewidth]{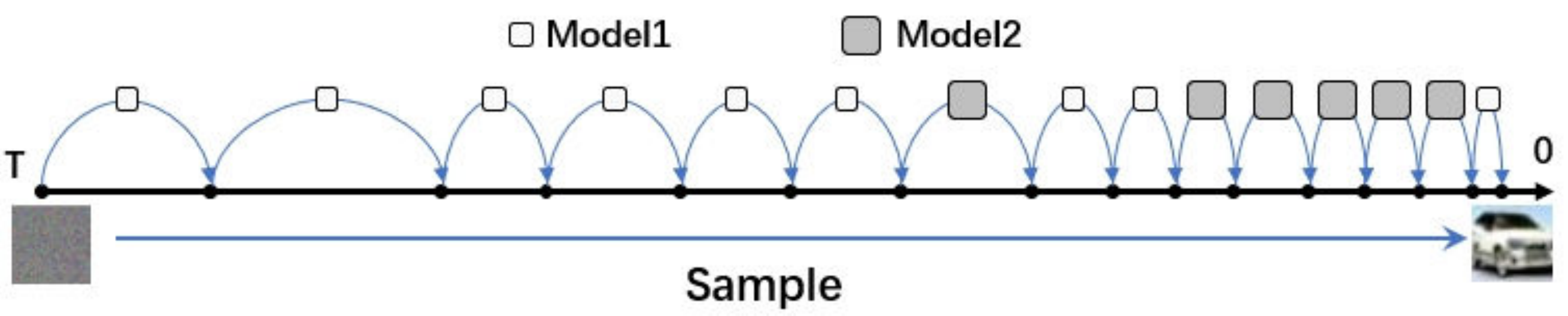}
    \caption{A searched schedule on CIFAR-10 using DDIM under 750ms budget.}
  \end{subfigure}
  \vfill
  \begin{subfigure}{0.8\linewidth}
    \centering
    \includegraphics[width=0.8\linewidth]{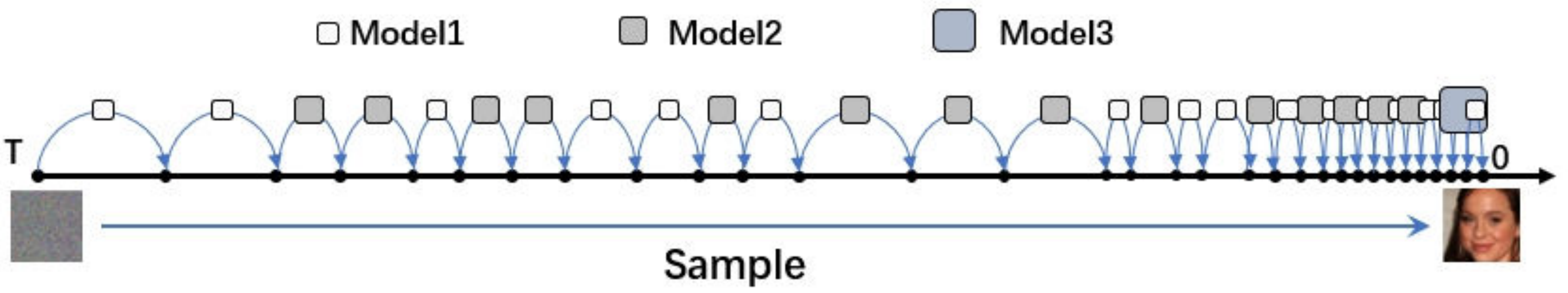}
    \caption{A searched schedule on CelebA using DDIM under 1500ms budget.}
  \end{subfigure}
  \vfill
  \begin{subfigure}{0.8\linewidth}
    \centering
    \includegraphics[width=0.8\linewidth]{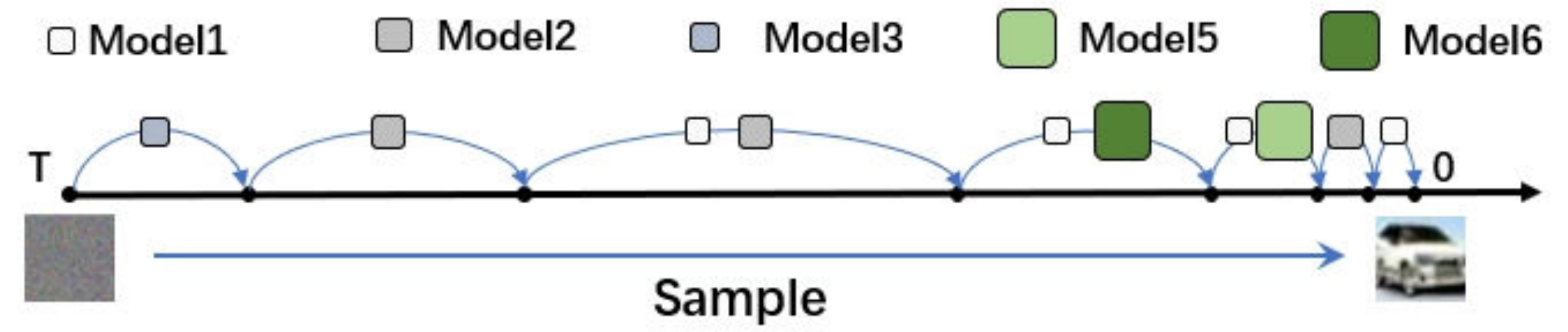}
    \caption{A searched schedule on CIFAR-10 using DPM-Solver under 700ms budget.}
  \end{subfigure}
  \vfill
  \begin{subfigure}{0.8\linewidth}
    \centering
    \includegraphics[width=0.8\linewidth]{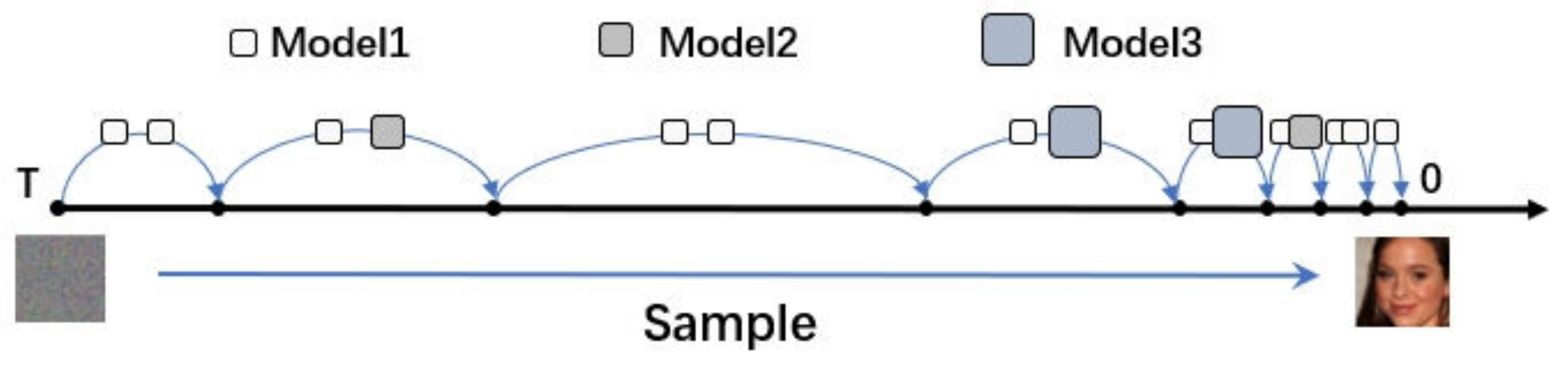}
    \caption{A searched schedule on CelebA using DPM-Solver under 650ms budget.}
  \end{subfigure}
  \vfill
  \begin{subfigure}{0.8\linewidth}
    \centering
    \includegraphics[width=0.8\linewidth]{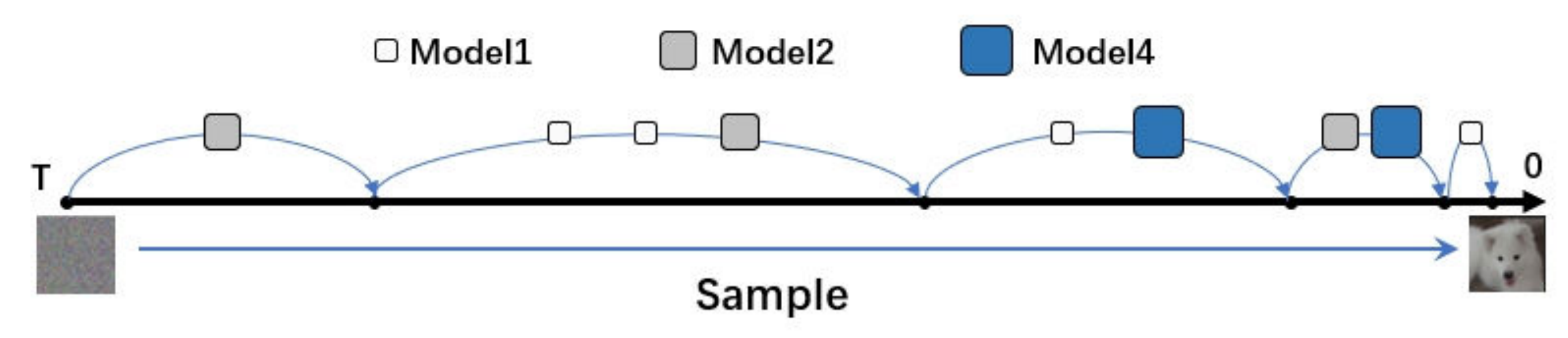}
    \caption{A searched schedule on ImageNet-64 using DPM-Solver under 800ms budget.}
  \end{subfigure}
  \vfill
  \begin{subfigure}{0.8\linewidth}
    \centering
    \includegraphics[width=0.8\linewidth]{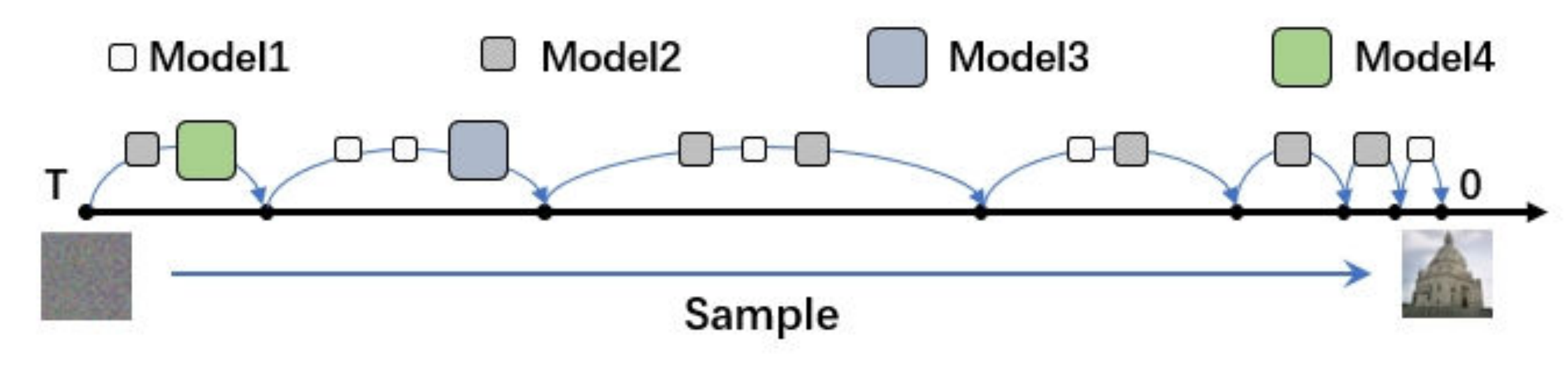}
    \caption{A searched schedule on LSUN-Church using DPM-Solver under 4000ms budget.}
  \end{subfigure}
  \caption{Several examples of searched schedules. The model numbers shown in the figure are consistent with the description at \cref{Sec:model zoo}. The sizes of model squares match the latencies of the corresponding models approximately.}
  \label{fig:search_pattern_full}
\end{figure*}

We show some searched schedules on all the four datasets using the two samplers at \cref{fig:search_pattern_full} (except for the searched patterns on ImageNet-64 and LSUN-Church using DDIM, which are shown at \cref{fig:search_pattern}).

\section{Model Zoo information}
\label{Sec:model zoo}

\subsection{Model Zoo Construction}
Leveraging a model zoo with varying complexities (models with different architectures) or functionalities (models with different architectures or training settings), our OMS-DPM method can derive model schedules that achieve superior speed-quality trade-offs. To obtain a model zoo, one can either construct and train the models themselves or directly use public models.

On CIFAR-10, CelebA, ImageNet-64, and LSUN-Church, in order to obtain models with varying complexities and functionalities,
we adjust the model architectures and train the models ourselves.
To be more specific, the most commonly used architecture family for DPMs is U-Net~\cite{unet,ddpm,sde,diffusion_beat_gan}.
A U-Net is constructed by a series of downsampling stages and upsampling stages. All of these blocks consist of several residual blocks followed by a downsampling or upsampling block. In addition, a global attention head is used after all residual blocks in some stages. 
And we adjust the U-Net architecture by changing the \textbf{(1)} widths of each stage (``Channels'' in \cref{tab:cifar10-model,tab:celeba-model,tab:imagenet64-model,tab:church-model}), \textbf{(2)} depth (``Depth'' in \cref{tab:cifar10-model,tab:celeba-model,tab:imagenet64-model,tab:church-model}), including the number of stages and the number of residual blocks per stage, and \textbf{(3)} the number of the attention head (``Attention'' in \cref{tab:cifar10-model,tab:celeba-model,tab:imagenet64-model,tab:church-model}). 
Then, we use the noise prediction form and linear noise schedule to train the models, and all other training settings are kept the same too. As the models have different architectures, they have different complexities and functionalities. In our experiments, the model zoo sizes are 6, 6, 7, and 6 on CIFAR-10, CelebA, ImageNet-64, and LSUN-Church, respectively. 

While in the experiment of using Stable Diffusion~\cite{ldm}, we use the four officially provided pre-trained models at \url{https://huggingface.co/CompVis/stable-diffusion} to construct the model zoo.

More detailed architecture configurations and training settings are illustrated as following.

\paragraph{CIFAR-10.} The architecture configurations are shown in \cref{tab:cifar10-model}. We choose the model 1,2,3,6 to construct the $N=4$ model zoo and model 2,6 to construct the $N=2$ model zoo in \cref{sec:ablation}. We follow the training settings in ~\cite{ddpm}. All models are trained with 128 batch size for 800k iterations, with a learning rate of $2\times10^{-4}$. We use 0.1 as dropout ratio and 0.9999 as EMA rate. 
\begin{table}[h]
  \centering
    \begin{tabular}{c|ccc}
      \toprule
      \textbf{Number} & \textbf{Channels} & \textbf{Depth} & \textbf{Attention} \\
      \midrule
      \textbf{1} & 32$\times$[1,2,2,2] & 2 & 1 head at 16$\times$16 \\
      \textbf{2} & 64$\times$[1,2,2,2] & 2 & 1 head at 16$\times$16 \\
      \textbf{3} & 128$\times$[1,1,1] & 1 & - \\
      \textbf{4} & 128$\times$[1,2,2,2] & 2 & - \\
      $\textbf{5}^{*}$ & 128$\times$[1,2,2,2] & 2 & 1 head at 16$\times$16 \\    
      \textbf{6} & 128$\times$[1,2,2,2] & 2 & 4 heads at 16$\times$16 \\
      \bottomrule
    \end{tabular}
    \caption{Architecture configuration on CIFAR-10. Model with * has the same architecture with the model used in ~\cite{ddpm}.}
    \label{tab:cifar10-model}
\end{table}

\paragraph{CelebA.} The architecture configurations are shown in Tab.~\ref{tab:celeba-model}. Following ~\cite{ddim}, we use a batch size of 128 and a learning rate of $2\times 10^{-4}$ for training. We train all models for 1000k iterations and save at every 50k, and then select the best among all checkpoints based on the FID with a 20-step DDIM sampler. We use 0.1 as dropout ratio and 0.9999 as EMA rate. 
\begin{table}[h]
    \centering
    \begin{tabular}{c|ccc}
      \toprule
      \textbf{Number} & \textbf{Channels} & \textbf{Depth} & \textbf{Attention} \\
      \midrule
      \textbf{1} & 32$\times$[1,2,2,2,4] & 2 & 1 head at 16$\times$16\\
      \textbf{2} & 64$\times$[1,2,2,2,4] & 2 & 1 head at 16$\times$16\\
      \textbf{3} & 64$\times$[1,2,2,2,4] & 3 & 1 head at 16$\times$16 \\
      $\textbf{4}^{*}$ & 128$\times$[1,2,2,2,4] & 2 & 1 head at 16$\times$16\\
      \textbf{5} & 160$\times$[1,2,2,2,4] & 2 & 1 head at 16$\times$16 \\
      \textbf{6} & 128$\times$[1,2,2,2,4] & 3 & 1 head at 16$\times$16 \\    
      \bottomrule
    \end{tabular}
    \caption{Architecture configuration on CelebA. Model with * has the same architecture with the model used in ~\cite{ddim}.}
    \label{tab:celeba-model}
\end{table}

\paragraph{ImageNet-64.} The architecture configurations are shown in Tab.~\ref{tab:imagenet64-model}. We follow the architecture choice in ~\cite{iddpm}, while we use our own settings for training. We set 128 as the batch size and $2\times10^{-5}$ as the learning rate. We do not train models to predict varience like ~\cite{iddpm} does. We choose the checkpoints with 1500k iterations. Dropout ratio 0.1 and EMA rate 0.9999 are used.
\begin{table}[t]
  \centering
  \begin{tabular}{c|ccc}
      \toprule
      \textbf{Number} & \textbf{Channels} & \textbf{Depth} & \textbf{Attention} \\
      \midrule
      \textbf{1} & 32$\times$[1,2,3,4] & 3 & 1 head at 16$\times$16 and 8$\times$8\\
      \textbf{2} & 64$\times$[1,2,3,4] & 3 & 1 head at 16$\times$16 and 8$\times$8\\
      \textbf{3} & 96$\times$[1,2,3,4] & 3 & 1 head at 16$\times$16 and 8$\times$8\\
      \textbf{4} & 128$\times$[1,2,3,4] & 2 & 1 head at 16$\times$16 and 8$\times$8\\  
      $\textbf{5}^{*}$ & 128$\times$[1,2,3,4] & 3 & 1 head at 16$\times$16 and 8$\times$8\\
      \textbf{6} & 128$\times$[1,2,3,4] & 4 & 1 head at 16$\times$16 and 8$\times$8\\
      \textbf{7} & 160$\times$[1,2,3,4] & 3 & 1 head at 16$\times$16 and 8$\times$8\\
      \bottomrule
  \end{tabular}
  \caption{Architecture configuration on ImageNet-64. Model with * has the same architecture with the model used in ~\cite{iddpm}.}
  \label{tab:imagenet64-model}
\end{table}
\paragraph{LSUN-Church.}  The architecture configurations are shown in \cref{tab:church-model}. We train these models with batch size 64 and learning rate $2\times10^{-5}$. We choose the checkpoints with 1000k iterations. Dropout ratio 0.1 and EMA rate 0.9999 are used.
\begin{table}[h]
    \centering
    \begin{tabular}{c|ccc}
      \toprule
      \textbf{Number} & \textbf{Channels} & \textbf{Depth} & \textbf{Attention} \\
      \midrule
      \textbf{1} & 32$\times$[1,1,2,2,4,4] & 2 & 1 heads at 16$\times$16\\
      \textbf{2} & 64$\times$[1,1,2,2,4,4] & 2 & 1 heads at 16$\times$16\\
      \textbf{3} & 96$\times$[1,1,2,2,4,4] & 2 & 1 heads at 16$\times$16 \\
      \textbf{4} & 128$\times$[1,1,2,2,4,4] & 1 & 1 heads at 16$\times$16 \\
      \textbf{5} & 128$\times$[1,1,2,2,4] & 1 & 1 heads at 32$\times$32 \\
      $\textbf{6}^{*}$ & 128$\times$[1,1,2,2,4,4] & 2 & 1 heads at 16$\times$16\\ 
      \bottomrule
    \end{tabular}
    \caption{Architecture configuration on LSUN-Church. Model with * has the same architecture with the model used in ~\cite{ddpm}.}
    \label{tab:church-model}
\end{table}

\paragraph{Stable Diffusion.}
The four models we use share the same architecture but are trained on different datasets and have different parameters. All four models are latent diffusion models that contain three modules: (1) A pre-trained CLIP text encoder is used to encode the prompt information. (2) A VAE maps high-dimension images into a low-dimension latent space. (3) A DPM conducts generation in the latent space using a U-Net guided by the prompt encoding. Note that the four models share the same VAE and CLIP text encoder, so we can safely compose the four U-Nets for the latent diffusion process.
\subsection{Inference Latency}
We test the time cost of inferring a batch of data for all models by averaging the latency over 500 inferences, and all inferences are conducted on a single A100 GPU. We list all results in Tab.~\ref{tab:latency}. The batch size $b$ is 64 for LSUN-Church, 128 for CelebA and ImageNet-64, and 512 for CIFAR-10. 
\begin{table}[h]
  \centering
  \begin{tabular}{c|cccc}
       \toprule
       \textbf{Number} & \textbf{CIFAR-10}(ms) & \textbf{CelebA}(ms) & \textbf{ImageNet-64}(ms) & \textbf{LSUN-Church}(ms) \\
       \midrule
       1 & 35.99$\pm$0.29 & 31.90$\pm$0.50 & 46.91$\pm$0.05 & 160.20$\pm$0.18 \\
       2 & 69.47$\pm$0.03 & 63.18$\pm$2.15 & 92.46$\pm$0.07 & 334.95$\pm$0.28 \\
       3 & 55.06$\pm$0.12 & 84.31$\pm$0.14 & 153.12$\pm$0.24 & 581.44$\pm$0.48 \\
       4 & 121.12$\pm$0.14 & 133.04$\pm$0.31 & 153.39$\pm$0.27 & 517.59$\pm$0.49 \\
       5 & 140.01$\pm$0.06 & 207.53$\pm$0.44 & 201.67$\pm$0.31 & 522.24$\pm$2.97 \\
       6 & 147.74$\pm$0.15 & 176.81$\pm$0.46 & 252.66$\pm$0.37 & 778.86$\pm$0.65\\
       7 & - & - & 309.93$\pm$1.42 & - \\    
       \bottomrule
  \end{tabular}
  \caption{Latency of all models in the model zoo.}
  \label{tab:latency}
\end{table}

\section{Experiment Details}
\label{Sec:experiment details}
\subsection{Evaluation of DPMs}
For the final evaluation of unconditional DPMs in \cref{tab:DPM-solver,tab:ddim,tab:ablation model zoo,tab:ablation training data}, we generate 50k images and calculate FID between the whole training set and the generated images. To reduce the randomness of evaluation for a fair comparison, we fix the 50k generation noise in all our experiments. Since we use ODE samplers, there is no randomness in our evaluation. For experiments with stable-diffusion, we sample 30k captions in the validation set and use them to guide the generation with a guidance scale $s=1.5$. Then we calculate the FID between generated images and the raw images of all sampled captions. 

\subsection{How to Conduct DPM Sampling for A Model Schedule}
\label{sec:app_sampler_ss_design}
In this section, we explain how we conduct the DPM sampling corresponding to a specific model schedule $\boldsymbol{q}=[s'_1,\cdots,s'_L]$.
\paragraph{DPM-Solver.} For DPM-Solver, a $k$-order DPM solver can be seen as grouping $k$ timesteps together. 
Our problem parametrization can still be used: we group every 3 timesteps to form $L/3$ groups. For example, $a_{s_1}, a_{s_2}, a_{s_3}$ belong to the first group and will be used together in a single solver step. Note that for each group, our parametrization also enables us to decide between inactive solver (e.g., $s_1=s_2=s_3=0$), first-order (e.g., $s_1\neq 0, s_2=s_3=0$), second-order (e.g., $s_1 \neq 0, s_2 \neq 0, s_3=0$), and third-order DPM solvers. Finally, we divide the continuous time following \emph{uniform logSNR} rule. The total number of time splits equals the number of used solver steps, or $L/3$ minus the number of inactive solvers.
Consider an example schedule [1,2,3,3,0,0,0,0,0,1,2,0] with length 3$\times$4. [1,2,3], [3,0,0], and [1,2,0] correspond to three solver steps, while [0,0,0] is not used during sampling. We split time [0,1] to four \emph{solver timesteps} (three time splits) with \emph{uniform logSNR}. We conduct a 2-nd order solver with model 2 and model 1 in order at the first split. Then we use a 1-st order solver with model 3 at the second split. Finally, we apply a 3-rd order solver using model 3, model 2, and model 1 in order at the last time split, and get the generated images.

We also adjust the sequence predictor to match the property of DPM-Solver. See \cref{app_sampler} for more details. For convenience, we set $L$ to be divisible by 3 in our experiments with DPM-Solver. 

For the maximum schedule length $L$, we set 90 on CIFAR-10 dataset, 60 on other datasets, and 45 for stable-diffusion.

\paragraph{DDIM.} Different from the experiments with DPM-Solver, the search space in our experiments with DDIM contains the time discretization scheme. Specifically, we linearly discretized [0,T] beforehand to get $L$ discrete timesteps. Then, the timesteps with non-zero $s'$, $\{t_i|s'_i \neq 0\}_{i=1,\cdots,L}$, are used in sampling, while the other timesteps $\{t_i|s'_i = 0\}_{i=1,\cdots,L}$ don't invovle in sampling. 
We set $L$ with DDIM as 200 on CelebA dataset and 100 on other datasets.

\subsection{Schedule-FID Data Generation}
\label{Sec:data form}
When generating schedule-FID data for predictor training, it's better to make the training data diversely distributed, such that the predictor can better generalize to unseen model schedules in the large search space. We set up multiple multinomial distributions by manually assigning the probabilities of picking each model from the model zoo. Then, for each multinomial distribution, we generate model schedules by sampling the model choice according to the distribution at each step. We will open source all the schedule-FID data for future use.
For each model schedule on all unconditional generation tasks, we use the corresponding DPM to sample 5k images and evaluate the FID score. For experiments with stable-diffusion, we randomly sample 1.5k captions from the MS COCO 256$\times$256 validation set for image generation, and then calculate the FID between generated images and the raw image of these sampled captions. Noting that the noise taken as input to generate schedule-FID data is also fixed.
\subsection{Adjustment of the Sequence Predictor for DPM-Solver}
\label{app_sampler}
DPM-Solver \cite{dpm-solver} uses a k-th solver to compute $x_{t_{i-1}}$ from $x_{t_i}$, which takes k NFE. So we make some adjustments to the sequence predictor module to match the characteristics of DPM-Solver. Specifically, we group each three model embeddings $[\mathrm{Emb}^\calM_{i-2}, \mathrm{Emb}^\calM_{i-1}, \mathrm{Emb}^\calM_{i}]$ ($i$ is divisible by 3), and then concatenate them to a 3$M$-dimension encoding. We feed this encoding into a MLP to get an encoding that represents the combination of three models, which we call solver embedding. Finally, the solver embedding is concatenated with timestep embedding and fed into the LSTM. 

\subsection{Hyperparameter of Predictor}
We set the model embedding dimension to 64 for ImageNet-64 as we use a larger model zoo size, and 32 for other datasets. For DPM-Solver, we set the solver embedding dimension to 64. For DDIM, we set the length of model embedding to 32. We used 64 as the dimension of timestep embedding. For the LSTM, we set the hidden size to 128 and the layer number to 1. Finally, we use an MLP with 4 layers and an output size of 200  at each layer except the last layer. No hyperparameter tuning is conducted.

\subsection{Training}
We use the ranking loss to train the predictor as described in \cref{sec:workflow}. When obtaining a batch of $b$ training data, we first randomly choose at most $compare\_ratio\times b$ pairs of data whose ground truth FID difference is larger than $threshold$. Then, we train the predictor with the ranking loss 
on these data. We set $compare\_ratio$ to 2, $threshold$ to 0.15 and the compare margin $m$ to 1.0.

\subsection{Validating the Reliability of Predictor}
We use Kendall's Tau (KD) to evaluate the performance of predictor \cite{kd} . Specifically, we split our generated schedule-FID dataset into two parts: training set and validation set. We complete the training procedure on the training set and test the KD between predicted scores and ground truth FID on the validation set. We report our results in \cref{tab:predictor}. And our predictor for stable-diffusion achieves a KD of 0.9543 on the validation set with 2070 training data and 108 validation data. Our predictors can achieve high KDs on unseen data, indicating their effectiveness.

\begin{table}[t]
  \centering
  \resizebox{0.5\textwidth}{!}{
  \begin{tabular}{c|c|ccc}
       \toprule
       \textbf{Sampler} & \textbf{Dataset} & \textbf{Train} & \textbf{Valid} & \textbf{KD} \\
       \midrule
       \multirow{4}{*}{\textbf{DPM-Solver}} & {CIFAR-10} & 3662 & 3662 & 0.9621 \\
       \cmidrule(lr){2-2}
       & {CelebA} & 3460 & 3460 & 0.9461 \\
       \cmidrule(lr){2-2}
       & {ImageNet-64} & 2735 & 2738 & 0.9611 \\
       \cmidrule(lr){2-2}
       & {LSUN-Church} & 3082 & 884 & 0.9283\\
       \midrule
       \multirow{4}{*}{\textbf{DDIM}} & {CIFAR-10} & 3380 & 3380 & 0.9757 \\
       \cmidrule(lr){2-2}
       & {CelebA} & 2660 & 2660 & 0.9653 \\
       \cmidrule(lr){2-2}
       & {ImageNet-64} & 3240 & 360 & 0.9760 \\
       \cmidrule(lr){2-2}
       & {LSUN-Church}  & 1838 & 94 & 0.9675 \\    
       \bottomrule
  \end{tabular}
  }
  \caption{Information of all predictors. Train/Valid means the total num of data in the training/validation sets. KD means the Kendall's Tau on validation set between predicted score and true FID.}
  \label{tab:predictor}
\end{table}

\subsection{Evolutionary Search}
\label{sec:app_evo}
Our complete search flow is shown at \cref{alg:evo}.
\begin{algorithm*}[tb]
    \caption{Predictor-based Evolutionary Search}
    \label{alg:evo}
    \begin{algorithmic}[1]
        \REQUIRE 
        ~\\$\mathrm{Perf}()$: a trained predictor
        ~\\$GetCost()$: a function to get the time cost of a model schedule by summing inference latency of all the models.
        
        \renewcommand{\algorithmicrequire}{\textbf{Input:}}
        \REQUIRE
        ~\\$\textbf{C}$: time budget of sampling a batch of images

        \renewcommand{\algorithmicrequire}{\textbf{Symbol:}}
        \REQUIRE
        ~\\$P$: The whole \emph{P}opulation of model schedule.
        ~\\$CP$: The \emph{C}andidate \emph{P}arents set of each loop, from which a parent model schedule is selected. 
        ~\\$NG$: The \emph{N}ext \emph{G}eneration newly mutated from the parent schedule in each loop.
        ~\\$E$: The \emph{E}liminated model schedules in each loop.

        \renewcommand{\algorithmicrequire}{\textbf{Hyperparameter:}}
        \REQUIRE
        ~\\$\textbf{Epoch}$: Number of loops for the entire search process.
        ~\\$\textbf{M}_{CP}$: Maximum size of the candidate parents set $CP$.
        ~\\$\textbf{iter}$: Maximum number of mutations in each loop.
        ~\\$\textbf{M}_{NG}$: Maximum size of the next generation set $NG$.
        ~\\$\textbf{M}_P$: Maximum size of the whole population $P$.
        
        \renewcommand{\algorithmicrequire}{\textbf{Search Process:}}
        \REQUIRE
        \STATE $P\gets\varnothing$
        \STATE Initialize a Schedule $\boldsymbol{q}_0$
        \STATE Add $\boldsymbol{q}_0$ to $P$
        \FOR{$t = 1, \cdots, \textbf{Epoch}$}
        \STATE $i=0$
        \STATE $NG\gets\varnothing$
        \STATE Random Sample $min(\textbf{M}_{CP},|P|)$ model schedules from $P$, denoted as $CP$
        \STATE Choose $\boldsymbol{q}$ with min $\mathrm{Perf}(\boldsymbol{q})$ in $CP$    
        \WHILE{$i<\textbf{itere}$ and $|NG|<\textbf{M}_{NG}$}    
        \STATE $\boldsymbol{q_{new}}\gets$Randomly mutate $\boldsymbol{q}$
        \IF{$GetCost(q_{new})<\textbf{C}$}
        \STATE add $\boldsymbol{q}_{new}$ to $NG$
        \ENDIF
        \STATE $i\gets i+1$
        \ENDWHILE
        \STATE $P\gets P\cup NG$
        \IF{$|P|>\textbf{M}_P$}
        \STATE Choose $|P|-\textbf{M}_P$ model schedules denoted as $E$ with max $\mathrm{Perf}(\boldsymbol{q}) (\boldsymbol{q}\in E)$ in $P$
        \STATE $P\gets P-E$
        \ENDIF
        \ENDFOR
    \end{algorithmic}
\end{algorithm*}
Time cost budget $\textbf{C}$ should be given in advance.
We first randomly initialize the whole schedule population $P$ with a single model schedule as described in lines 1-3. To ensure the initial schedule $\boldsymbol{q}_0$ falls in a region with good quality, we ensure that it has a time cost in [0.9$\times\textbf{C}$, $\textbf{C}$]. Then we conduct $T$ loops, in each of which we sample a parent schedule from the current $P$ and mutate the parent schedule to get a new candidate schedule. Specifically, we randomly sample at most $\textbf{M}_{CP}$ schedules in the $P$ as candidate parent set (denoted as $CP$) and choose the best one (denoted as $\boldsymbol{q}$ in line 8) as parent according to the predicted score. Then we mutate the parent to get more schedules denoted as $\boldsymbol{q_{new}}$ and add them into the next generation set (denoted as $NG$). The mutation is conducted for at most $\textbf{iter}$ times or until the size of $NG$ reaches $\textbf{M}_{NG}$. Then all schedules in $NG$ are added to the current population. If the size of $P$ is more than $\textbf{M}_P$, we eliminate excess according to the predicted score, as described in lines 18-19. Finally, after $T$ loops, we choose the one with the best predicted score in the population as our searched schedule.

 We set the maximum times $\textbf{iter}$ of random mutation as 200 and the maximum population $\textbf{M}_{NG}$ of the next generation as 40. The number $\textbf{p}$ of candidate parents at every epoch is 10. The population cap $\textbf{M}_P$ is set to 40. We set the total search epoch T as 600, but according to our experience, in most cases the search can reach the local optimum and be terminated around 
 200$\sim$300 epochs.


\subsection{\update{Implementation Details and Full Results of Baseline (1)}}
\label{Sec:details baseline1}
\update{
We have reported the \textbf{best} FID achieved by DPMs using a single model and several common sampler settings (baseline \textbf{(1)}) under every budget in \cref{ddim_dpm_experiments}. The complete results of the generation quality are shown in \cref{tab:dpm+full} and \cref{tab:ddim+full}.
The full results support many of our analysis in \cref{sec:empirical_obs}. For example, smaller models converge more quickly at relatively lower budgets with worse convergence generation quality, and \emph{quadratic} time discrete scheme is significantly better than \emph{linear} time discrete scheme on CIFAR-10, CelebA and ImageNet-64 while the case is just the opposite on LSUN-Church.
While the optimal hyper-parameters (e.g., the order of DPM-Solver, the time discretization scheme, the model size) are different for different datasets, our OMS-DPM can always outperform the baselines. This suggests the effectiveness of OMS-DPM in automatically finding good hyper-parameters and the benefits of reducing the burden of manual hyper-parameter tuning.
\par
For DPM-Solver \cite{dpm-solver}, we apply the fast version for 1-st, 2-nd and 3-rd order solver without adaptive step size based on the official implementation at \url{https://github.com/LuChengTHU/dpm-solver}. We apply \emph{uniform logSNR} time discrete scheme following the default configuration.\par
For DDIM \cite{ddim}, we obtain the results by using \emph{quadratic} time discrete scheme and \emph{uniform} time discrete scheme on all four datasets, following the official implementation at \url{GitHub - ermongroup/ddim: Denoising Diffusion Implicit Models}.\par
For stable-diffusion, we choose the single-step DPM-Solver and apply the \emph{uniform logSNR} time discrete scheme. We also apply the fast version for 1-st, 2-nd and 3-rd order solver and choose the best one. Other settings are kept consistent with the default configuration of the official implementation at \url{GitHub - CompVis/stable-diffusion: A latent text-to-image diffusion model}.
}
\begin{table}[t]
\begin{subtable}[t]{1.0\linewidth}
\centering
\resizebox{1.0\textwidth}{!}{
    \begin{tabular}{c|cccccc}
      \toprule
      \multirow{2}{*}{\textbf{Budget/ms}} & \multicolumn{6}{c}{\textbf{Model Number}} \\
      \cmidrule(lr){2-7} 
       & \textbf{1} & \textbf{2} & \textbf{3} & \textbf{4} & \textbf{5} & \textbf{6} \\
      \midrule
      $7.0\times10^3$ & 15.89/15.54/15.54 & 10.62/9.83/9.84 & 7.56/6.59/6.57 & 6.36/4.65/4.62 & 5.24/3.74/3.70 & 5.29/3.57/\textbf{3.56}\\
      $4.0\times10^3$ & 16.23/15.53/15.56 & 8.49/6.60/6.60 & 11.36/9.83/9.83 & 8.09/4.72/4.56 & 7.13/3.84/3.71 & 7.38/3.63/\textbf{3.61}\\
      $2.5\times10^3$ & 16.77/15.49/15.51 & 10.19/6.65/6.56 & 12.58/9.84/9.84 & 11.44/4.81/4.64 & 10.74/4.17/4.07 & 11.09/3.95/\textbf{3.93}\\
      $1.4\times10^3$ & 18.26/15.46/15.51 & 14.11/6.92/6.81 & 15.72/10.01/9.75 & 22.57/8.09/\textbf{5.23} & 22.95/6.70/6.01 & 24.25/7.41/6.89 \\
      $0.7\times10^3$& 22.49/15.69/15.74 & 28.68/\textbf{8.73}/12.22 & 27.59/11.44/10.10 & 50.29/39.04/23.07 & 60.33/69.40/289.53 & 56.55/43.14/297.67 \\
      \bottomrule
    \end{tabular}
    }
\label{tab:dpm+CIFAR-10+full}
\caption{Full results on CIFAR-10}
\end{subtable}

\begin{subtable}[t]{1.0\linewidth}
\centering
\resizebox{1.0\textwidth}{!}{
    \begin{tabular}{c|cccccc}
      \toprule
      \multirow{2}{*}{\textbf{Budget/ms}} & \multicolumn{6}{c}{\textbf{Model Number}} \\
      \cmidrule(lr){2-7} 
       & \textbf{1} & \textbf{2} & \textbf{3} & \textbf{4} & \textbf{5} & \textbf{6} \\
      \midrule
      $7.0\times10^3$ & 8.79/8.66/8.67 & 4.07/3.30/3.31 & 3.63/2.82/2.81 & 5.30/3.55/3.52 & 4.57/2.73/2.53 & 4.56/2.53/\textbf{2.49}\\
      $5.0\times10^3$ & 8.86/8.64/8.66 & 4.41/3.30/3.32 & 4.05/2.83/2.80 & 6.41/3.56/3.54 & 5.54/2.80/2.50 & 5.55/2.60/\textbf{2.49}\\
      $3.0\times10^3$ & 9.70/8.64/8.66 & 5.31/3.29/3.30 & 5.10/2.84/2.77 & 8.32/3.52/3.60 & 8.50/3.68/2.76 & 8.51/2.92/\textbf{2.40}\\
      $1.5\times10^3$ & 9.81/8.52/8.70 & 14.11/6.92/6.81 & 8.34/3.24/\textbf{2.78} & 17.59/4.53/6.63 & 22.34/9.13/79.20 & 18.30/5.04/10.83 \\
      $0.65\times10^3$& 9.81/8.52/8.70 & 18.44/\textbf{4.79}/6.84 & 22.26/10.19/80.42 & 45.06/151.41/352.54 & 42.44/316.57/333.59 & 43.94/305.22/314.81 \\
      \bottomrule
    \end{tabular}
    }
\label{tab:dpm+CelebA+full}
\caption{Full results on CelebA}
\end{subtable}

\begin{subtable}[t]{1.0\linewidth}
\centering
\resizebox{1.0\textwidth}{!}{
    \begin{tabular}{c|ccccccc}
      \toprule
      \multirow{2}{*}{\textbf{Budget/ms}} & \multicolumn{6}{c}{\textbf{Model Number}} \\
      \cmidrule(lr){2-8} 
       & \textbf{1} & \textbf{2} & \textbf{3} & \textbf{4} & \textbf{5} & \textbf{6} & \textbf{7}\\
      \midrule
      $12\times10^3$ & 40.73/40.36/40.66 & 23.75/23.30/23.67 & 18.67/17.86/18.13 & 17.40/16.58/16.89 & 15.89/14.79/14.94 & 15.77/1404/14.10 & 15.43/\textbf{12.99}/13.04\\
      $8.0\times10^3$ & 40.95/40.34/40.64 & 24.16/23.41/23.68 & 19.54/18.10/18.23 & 18.25/16.79/16.90 & 17.43/15.10/14.97 & 17.95/14.49/14.34 & 18.66/13.67/\textbf{13.44}\\
      $5.0\times10^3$ & 41.40/40.39/40.59 & 25.18/23.59/23.67 & 21.97/18.53/18.24 & 20.53/17.20/17.00 & 21.23/15.89/15.29 & 23.23/15.66/14.76 & 25.25/15.19/\textbf{14.00}\\
      $2.0\times10^3$ & 43.87/40.65/40.62 & 31.74/25.26/23.96 & 36.74/21.79/19.51 & 35.99/20.50/\textbf{18.20} & 46.09/23.92/20.63 & 57.25/30.84/33.40 & 66.38/30.80/41.10 \\
      $0.8\times10^3$& 52.35/41.79/42.59 & 59.70/33.86/\textbf{29.59} & 81.26/52.18/42.97 & 82.16/57.88/39.99 & 120.38/276.14/254.11 & 118.21/250.15/243.33 & 162.16/209.78/209.77 \\
      \bottomrule
    \end{tabular}
    }
\label{tab:dpm+ImageNet64+full}
\caption{Full results on ImageNet-64}
\end{subtable}

\begin{subtable}[t]{1.0\linewidth}
\centering
\resizebox{1.0\textwidth}{!}{
    \begin{tabular}{c|cccccc}
      \toprule
      \multirow{2}{*}{\textbf{Budget/ms}} & \multicolumn{6}{c}{\textbf{Model Number}} \\
      \cmidrule(lr){2-7} 
       & \textbf{1} & \textbf{2} & \textbf{3} & \textbf{4} & \textbf{5} & \textbf{6} \\
      \midrule
      $35\times10^3$ & 134.10/133.72/133.71 & 56.47/54.58/54.51 & 18.89/16.30/16.16 & 17.46/15.58/15.40 & 19.72/17.42/17.19 & 15.20/12.36/\textbf{11.97}\\
      $25\times10^3$ & 134.25/133.58/133.72 & 57.55/54.58/54.51 & 20.32/16.43/16.12 & 18.53/15.72/15.43 & 20.86/17.47/17.20 & 16.85/12.55/\textbf{12.02}\\
      $15\times10^3$ & 134.78/133.21/133.43 & 59.89/54.43/55.08 & 24.20/17.33/15.91 & 21.43/16.11/15.65 & 24.29/18.13/16.77 & 21.64/13.54/\textbf{12.03}\\
      $10\times10^3$ & 135.63/132.99/132.99 & 63.56/55.02/53.88 & 29.64/18.93/17.08 & 25.63/16.70/15.57 & 28.68/18.98/16.92 & 26.86/14.24/\textbf{13.23} \\
      $4.0\times10^3$& 141.11/128.36/129.89 & 79.94/60.46/59.48 & 63.64/32.95/49.83 & 48.21/\textbf{32.23}/73.31 & 53.22/35.54/67.31 & 65.04/123.53/105.1 \\
      \bottomrule
    \end{tabular}
    }
\label{tab:dpm+Church+full}
\caption{Full results on LSUN-Church}
\end{subtable}
\caption{Complete FID results of baseline (1) using 1-st/2-nd/3-rd order of DPM-Solver on four datasets.}
\label{tab:dpm+full}
\end{table}

\begin{table}[t]
\begin{subtable}[t]{1.0\linewidth}
\centering
\resizebox{1.0\textwidth}{!}{
    \begin{tabular}{c|cccccc}
      \toprule
      \multirow{2}{*}{\textbf{Budget/ms}} & \multicolumn{6}{c}{\textbf{Model Number}} \\
      \cmidrule(lr){2-7} 
       & \textbf{1} & \textbf{2} & \textbf{3} & \textbf{4} & \textbf{5} & \textbf{6} \\
      \midrule
      $9.0\times10^3$ & 15.14/14.91 & 6.56/7.17 & 10.17/10.74 & 4.77/5.34 & 4.86/4.52 & 4.38/\textbf{4.29} \\
      $6.0\times10^3$ & 15.00/15.01 & 6.60/7.38 & 10.33/10.93 & 5.16/5.76 & 5.54/4.93 & 5.04/\textbf{4.73}\\
      $3.0\times10^3$ & 14.75/15.30 & 7.11/8.12 & 11.02/11.54 & 7.29/7.32 & 8.45/\textbf{6.42} & 7.86/6.47\\
      $1.5\times10^3$ & 14.82/16.00 & 9.57/\textbf{10.01} & 13.37/13.06 & 13.08/11.57 & 15.54/11.59 & 14.60/11.73  \\
      $0.75\times10^3$& 16.73/17.95 & 16.81/\textbf{16.11} & 20.00/17.45 & 26.36/24.31 & 30.08/27.02 & 29.91/27.91 \\
      \bottomrule
    \end{tabular}
    }
\label{tab:ddim+CIFAR-10+full}
\caption{Full results on CIFAR-10}
\end{subtable}

\begin{subtable}[t]{1.0\linewidth}
\centering
\resizebox{1.0\textwidth}{!}{
    \begin{tabular}{c|cccccc}
      \toprule
      \multirow{2}{*}{\textbf{Budget/ms}} & \multicolumn{6}{c}{\textbf{Model Number}} \\
      \cmidrule(lr){2-7} 
       & \textbf{1} & \textbf{2} & \textbf{3} & \textbf{4} & \textbf{5} & \textbf{6} \\
      \midrule
      $15\times10^3$ & 8.44/9.79 & 4.74/5.57 & 4.63/5.19 & 6.48/4.98 & 6.51/\textbf{4.61} & 6.48/4.88\\
      $10\times10^3$ & 8.46/9.81 & 5.42/5.62 & 5.53/5.25 & 7.78/5.10 & 8.04/\textbf{4.75} & 7.98/5.04\\
      $7.0\times10^3$ & 8.57/9.84 & 6.36/5.69 & 6.54/5.33 & 9.08/5.30 & 9.71/\textbf{5.03} & 9.49/5.37\\
      $4.0\times10^3$ & 8.92/9.85 & 8.22/5.89 & 8.88/\textbf{5.64} & 11.48/5.97 & 12.28/5.92 & 11.80/6.30 \\
      $1.5\times10^3$& 10.74/9.97 & 13.12/\textbf{7.32} & 13.90/7.75 & 15.61/9.65 & 17.07/11.93 & 15.84/12.83 \\
      \bottomrule
    \end{tabular}
    }
\label{tab:ddim+CelebA+full}
\caption{Full results on CelebA}
\end{subtable}

\begin{subtable}[t]{1.0\linewidth}
\centering
\resizebox{1.0\textwidth}{!}{
    \begin{tabular}{c|ccccccc}
      \toprule
      \multirow{2}{*}{\textbf{Budget/ms}} & \multicolumn{6}{c}{\textbf{Model Number}} \\
      \cmidrule(lr){2-8} 
       & \textbf{1} & \textbf{2} & \textbf{3} & \textbf{4} & \textbf{5} & \textbf{6} & \textbf{7}\\
      \midrule
      $20\times10^3$ & 41.91/42.29 & 24.41/25.71 & 19.18/19.82 & 18.11/18.66 & 16.57/16.61 & 16.05/15.71 & 15.08/\textbf{14.74}\\
      $15\times10^3$ & 42.07/42.31 & 24.64/25.78 & 19.53/19.96 & 18.46/18.78 & 16.95/16.81 & 16.55/16.04  & 15.67/\textbf{15.12}\\
      $10\times10^3$ & 42.44/42.40 & 25.01/25.96 & 20.12/20.32 & 19.53/19.17 & 17.71/17.29 & 17.53/16.74 & 16.81/\textbf{16.12} \\
      $5.0\times10^3$ & 43.36/42.72 & 26.05/26.61 & 21.73/21.78 & 20.76/20.50 & 20.12/\textbf{19.47} & 20.66/19.95 & 20.78/20.13 \\
      $2.0\times10^3$& 45.49/44.06 & 29.13/29.74 & 27.56/28.34 & \textbf{26.91}/27.00 & 30.77/31.38 & 36.25/37.36 & 40.09/41.49 \\
      \bottomrule
    \end{tabular}
    }
\label{tab:ddim+ImageNet64+full}
\caption{Full results on ImageNet-64}
\end{subtable}

\begin{subtable}[t]{1.0\linewidth}
\centering
\resizebox{1.0\textwidth}{!}{
    \begin{tabular}{c|cccccc}
      \toprule
      \multirow{2}{*}{\textbf{Budget/ms}} & \multicolumn{6}{c}{\textbf{Model Number}} \\
      \cmidrule(lr){2-7} 
       & \textbf{1} & \textbf{2} & \textbf{3} & \textbf{4} & \textbf{5} & \textbf{6} \\
      \midrule
      $55\times10^3$ & 134.62/52.35 & 51.32/38.01 & 15.28/31.86 & 14.71/30.11 & 16.43/32.84 & \textbf{11.72}/26.44\\
      $40\times10^3$ & 133.46/52.54 & 49.85/38.37 & 15.08/32.40 & 14.53/30.86 & 16.36/33.41 & \textbf{11.73}/27.00\\
      $25\times10^3$ & 129.85/52.87 & 46.22/39.07 & 14.71/33.63 & 14.38/31.64 & 16.14/34.77 & \textbf{12.05}/28.27\\
      $10\times10^3$ & 119.75/54.20 & 35.45/42.64 & 15.58/39.44 & \textbf{14.77}/35.81 & 16.72/39.68 & 15.48/35.17 \\
      $4.0\times10^3$& 97.27/57.64 & 28.07/51.87 & 30.93/60.99 & \textbf{25.27}/51.62 & 28.14/56.71 & 33.85/58.05 \\
      \bottomrule
    \end{tabular}
    }
\label{tab:ddim+Church+full}
\caption{Full results on LSUN-Church}
\end{subtable}
\caption{Complete FID results of baseline (1) using linear/quadratic time discretization scheme on four datasets with DDIM.}
\label{tab:ddim+full}
\end{table}

\section{Generated Images}
\begin{figure}[h]
  \centering
  \begin{subfigure}{0.45\linewidth}
    \centering
    \includegraphics[width=0.9\linewidth]{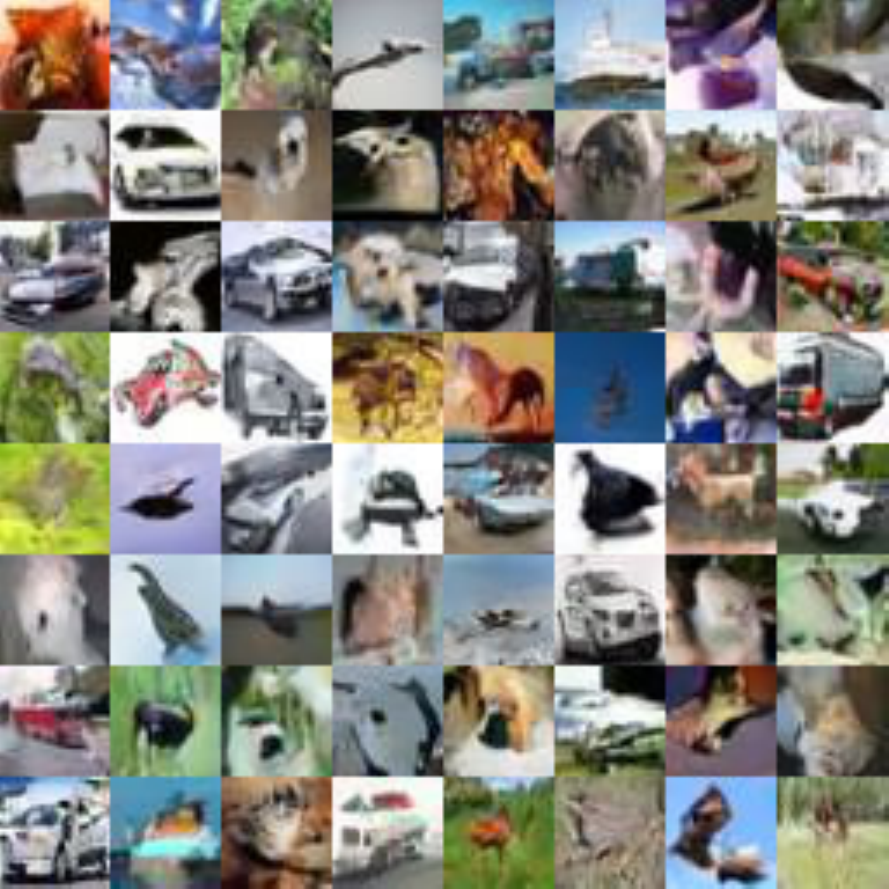}
    \caption{Samples generated by the baseline (1) method. FID=8.73}
  \end{subfigure}
  \hfill
  \begin{subfigure}{0.45\linewidth}
    \centering
    \includegraphics[width=0.9\linewidth]{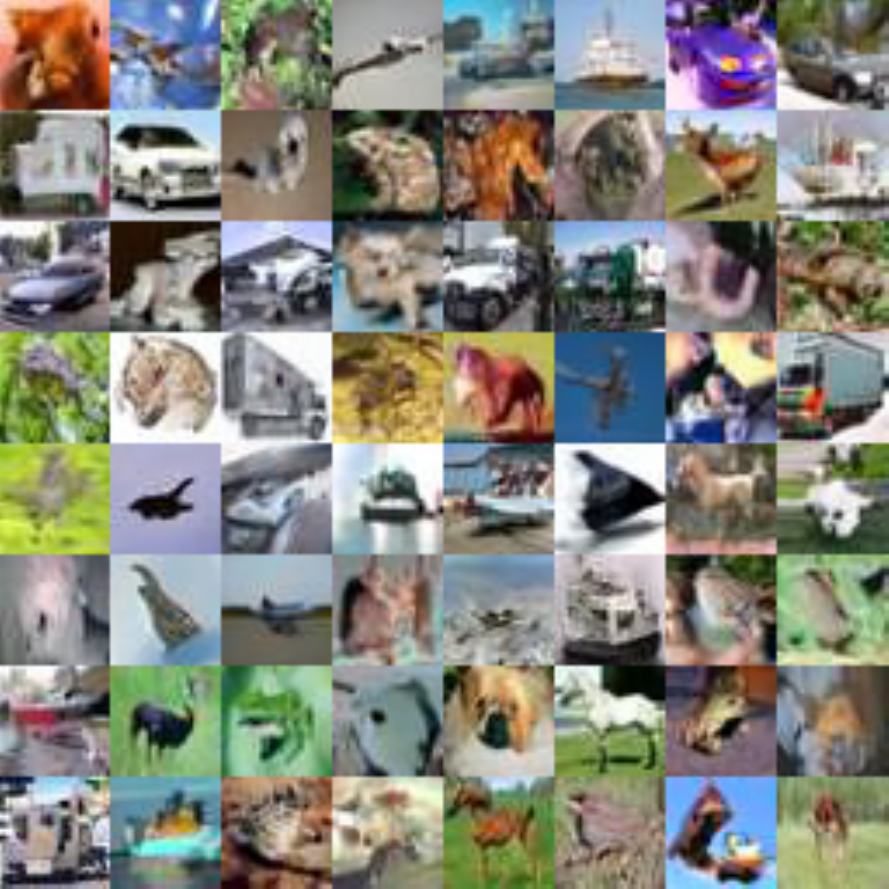}
    \caption{Samples generated by the \nameshort{}. FID=6.08}
  \end{subfigure}
  \caption{Samples of CIFAR-10 dataset under 700ms budget.}
  \label{fig:samples_CIFAR-10}
\end{figure}

\begin{figure}[h]
  \centering
  \begin{subfigure}{0.45\linewidth}
    \centering
    \includegraphics[width=0.9\linewidth]{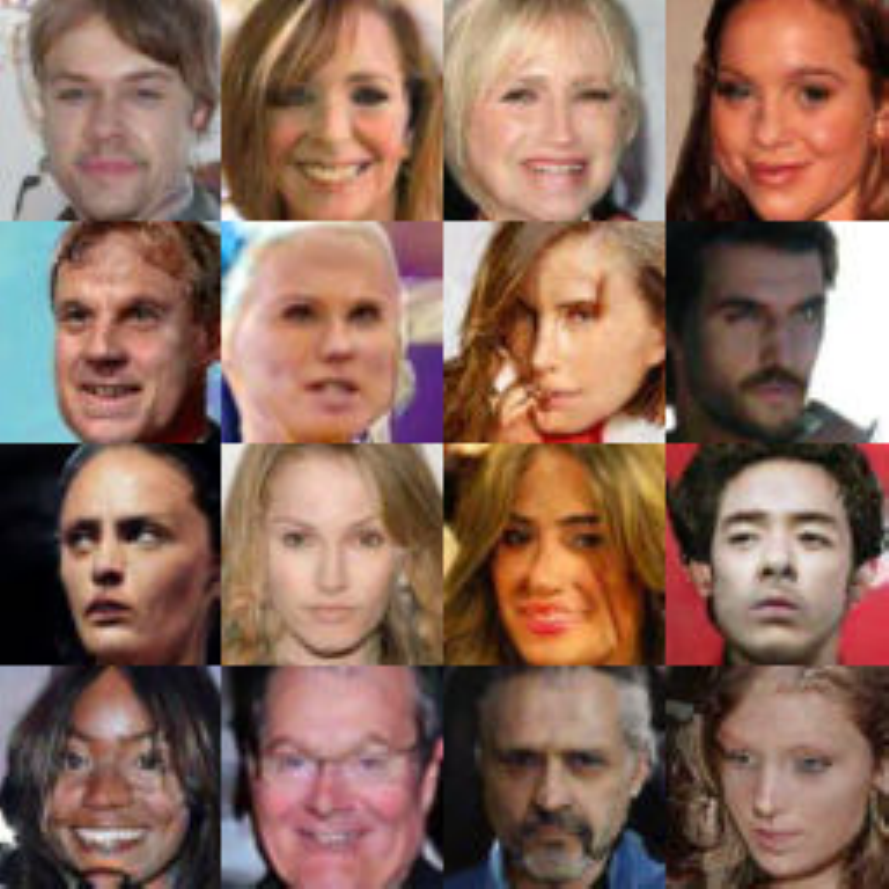}
    \caption{Samples generated by the baseline (1) method. FID=4.79}
  \end{subfigure}
  \hfill
  \begin{subfigure}{0.45\linewidth}
    \centering
    \includegraphics[width=0.9\linewidth]{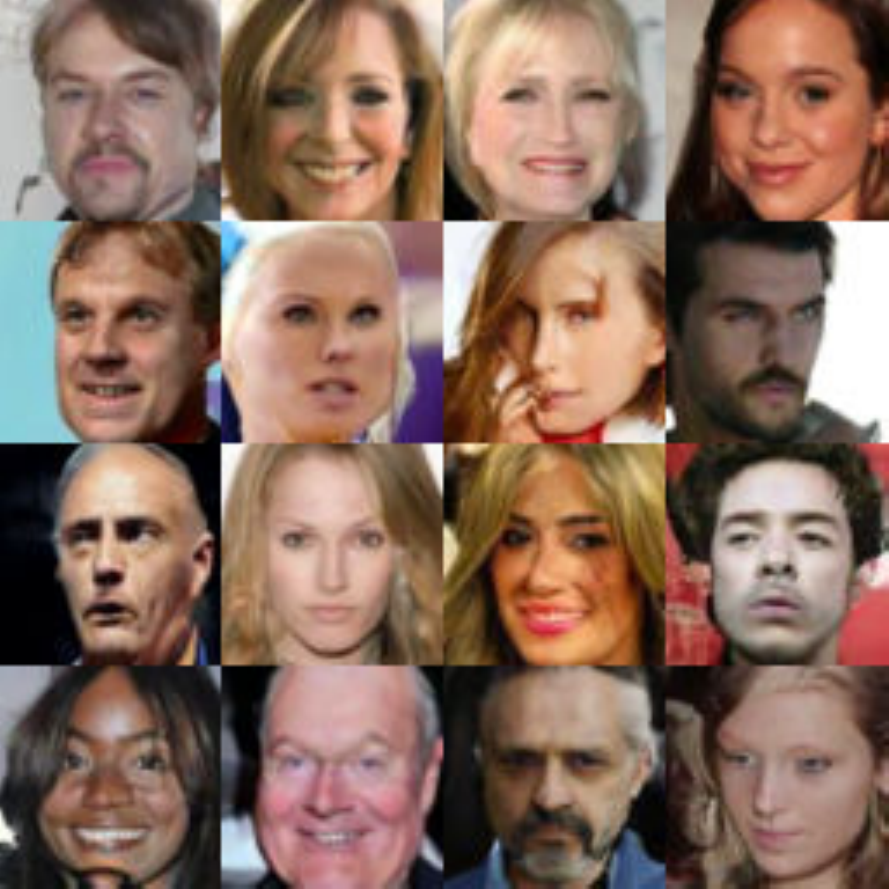}
    \caption{Samples generated by the \nameshort{}. FID=3.53}
  \end{subfigure}
  \caption{Samples of CelebA dataset under 650ms budget.}
  \label{fig:samples_CelebA}
\end{figure}

\begin{figure}[h]
  \centering
  \begin{subfigure}{0.45\linewidth}
    \centering
    \includegraphics[width=0.9\linewidth]{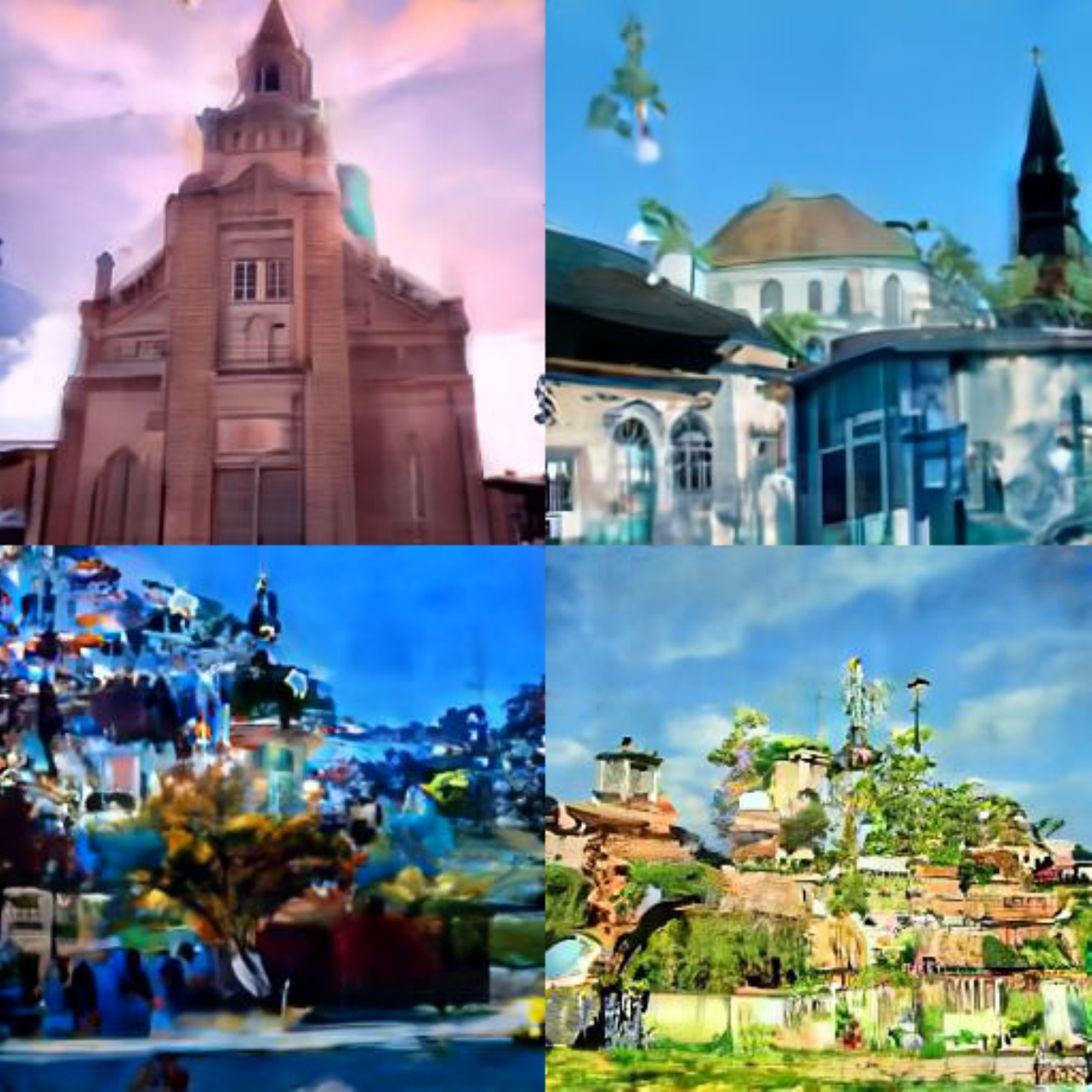}
    \caption{Samples generated by the baseline (1) method. FID=32.23}
  \end{subfigure}
  \hfill
  \begin{subfigure}{0.45\linewidth}
    \centering
    \includegraphics[width=0.9\linewidth]{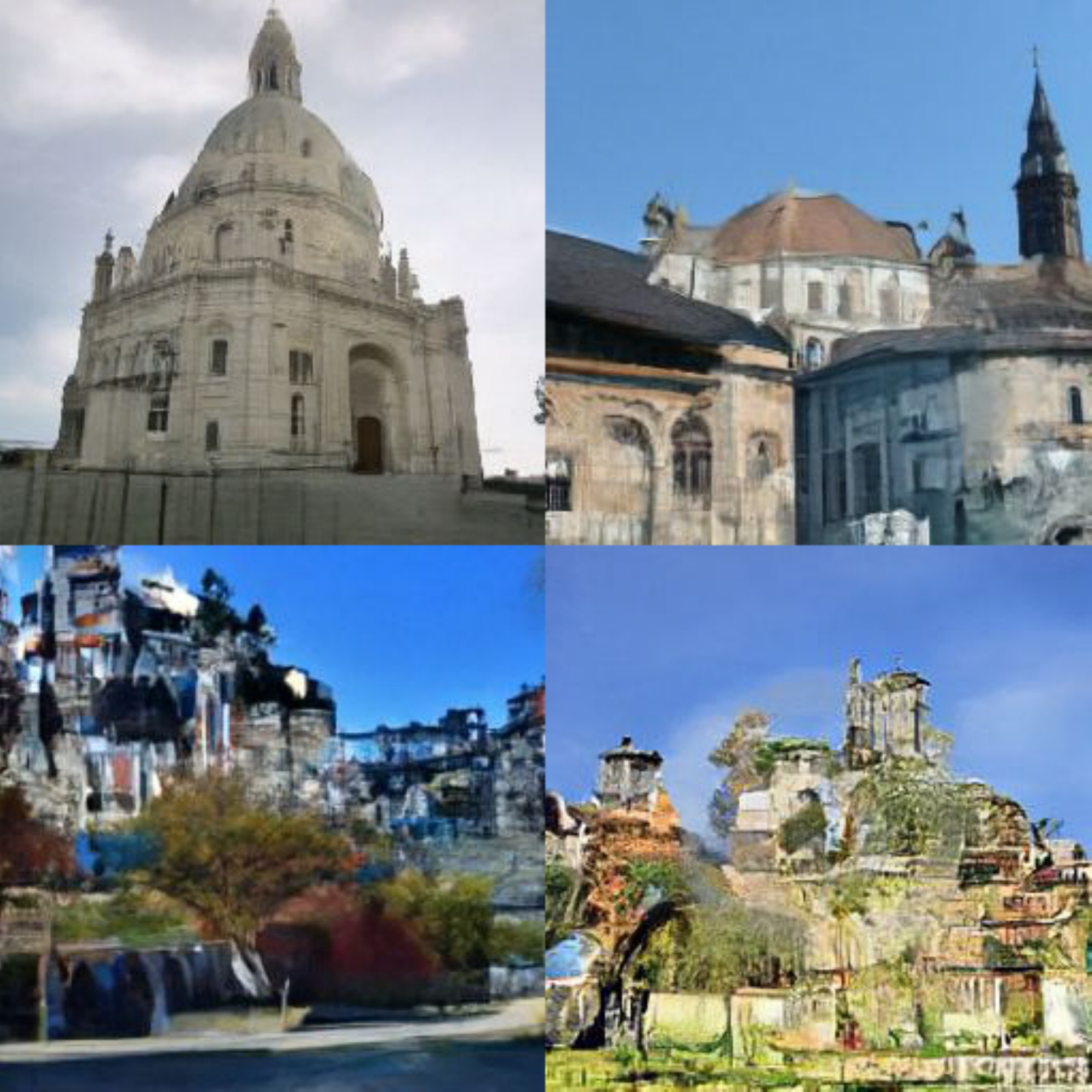}
    \caption{Samples generated by the \nameshort{}. FID=13.94}
  \end{subfigure}
  \caption{Samples of LSUN-Church dataset under 4000ms budget.}
  \label{fig:samples_Church}
\end{figure}

\begin{figure}[h]
  \centering
  \begin{subfigure}{0.45\linewidth}
    \centering
    \includegraphics[width=0.9\linewidth]{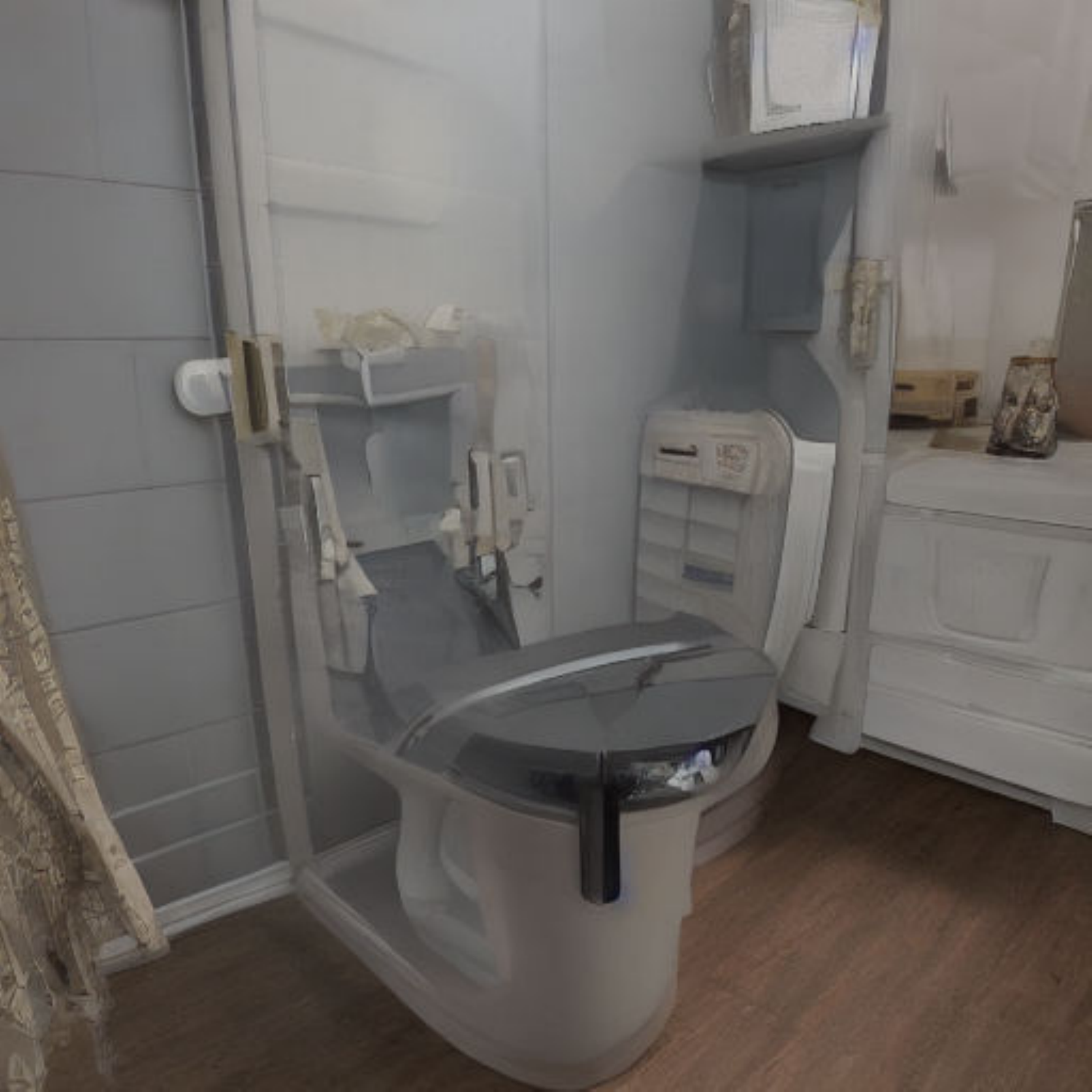}
    \caption{Samples generated by the baseline (1) method. FID=11.92}
  \end{subfigure}
  \hfill
  \begin{subfigure}{0.45\linewidth}
    \centering
    \includegraphics[width=0.9\linewidth]{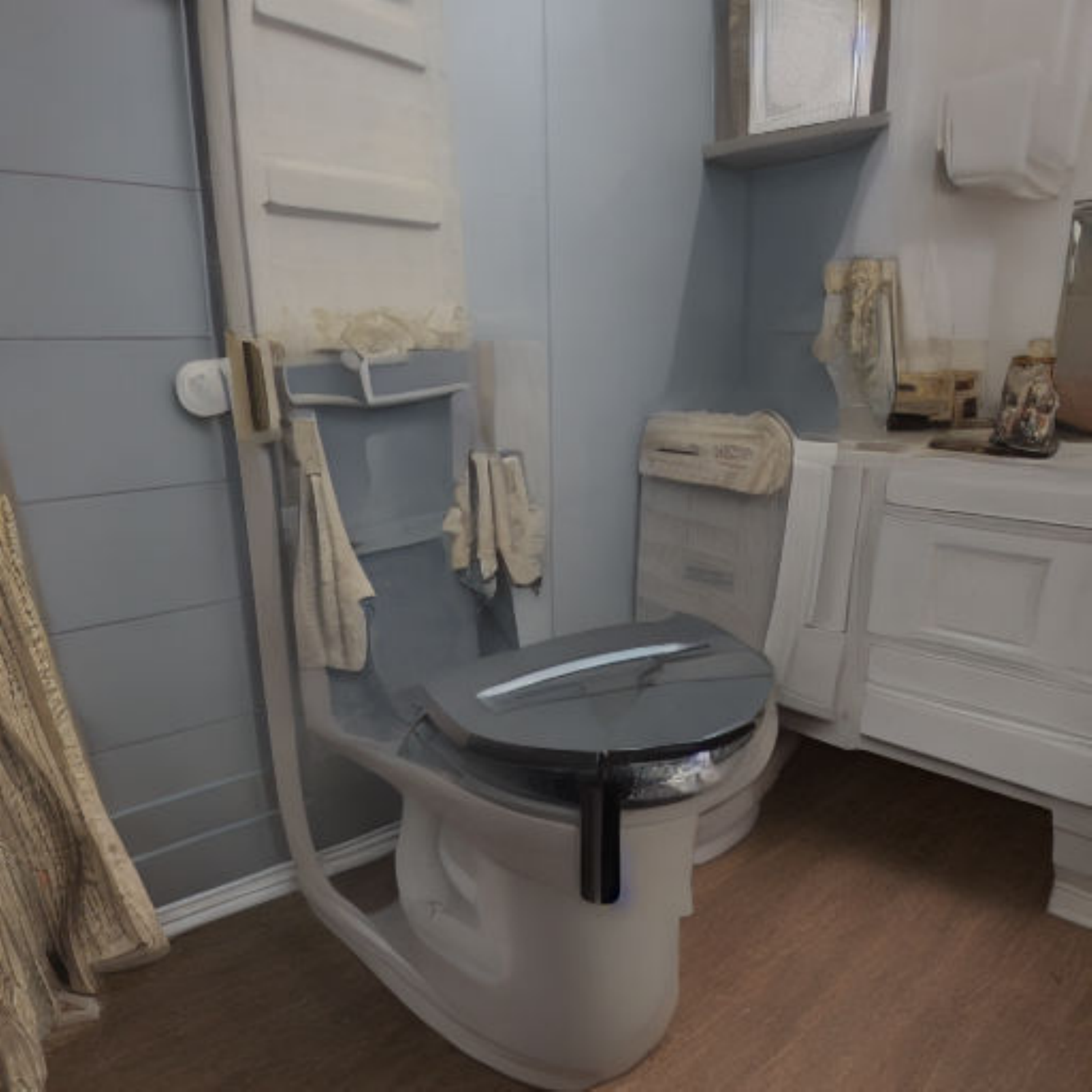}
    \caption{Samples generated by the \nameshort{}. FID=10.72}
  \end{subfigure}
  \caption{Sample of MS-COCO 256$\times$256 under 15 NFE budget, guided by the caption ``A small closed toilet in a cramped space''.}
  \label{fig:samples_coco_2}
\end{figure}

We put some samples using our method and baseline (1) under the lowest budget with DPM-Solver in this section at \cref{fig:samples_CIFAR-10}, \cref{fig:samples_CelebA}, \cref{fig:samples_Church} and \cref{fig:samples_coco_2}.

\end{document}